\documentclass{article}

\usepackage[table,xcdraw]{xcolor}
\usepackage[preprint]{corl_2026} 
\usepackage{multirow}
\usepackage{amssymb}
\usepackage{float}
\usepackage{subcaption}
\usepackage{graphicx}
\usepackage{natbib}
\usepackage{amsmath}
\usepackage{caption}
\usepackage{cleveref} 
\usepackage{enumitem}
\usepackage[all]{nowidow}
\usepackage{soul}
\usepackage{caption}
\usepackage{tcolorbox}
\tcbuselibrary{breakable}
\usepackage{fancyvrb}   
\newcounter{mysubsubsec}[subsection]
\renewcommand{\themysubsubsec}{\thesubsection.\arabic{mysubsubsec}}
\crefname{mysubsubsec}{section}{sections}  
\Crefname{mysubsubsec}{Section}{Sections}

\newcommand{\numberedparagraph}[1]{%
  \refstepcounter{mysubsubsec}%
  \noindent\textbf{\themysubsubsec\enspace #1}%
}

\usepackage{wrapfig}

\newcommand{\METHOD}{SeededGrasp}

\usepackage{uoftcolors}

\definecolor{rowblue}{RGB}{199, 224, 235}

\title{SeededGrasp: Language-Guided Grasping in Complex Scenes with Multiple Embodiments}

\author{
    Yang Xu$^{1,2,}$\thanks{Shared first.}, \quad 
    Gurpreet Singh Mukker$^{1,}$\footnotemark[1], \quad 
    Raymond Wang$^{3,}$\footnotemark[1] \\
    \textbf{Jasper Gerigk}$^{1,2}\textbf{,}$ \quad
    \textbf{Maria Attarian}$^{1,2,4}\textbf{,}$ \quad 
    \textbf{Igor Gilitschenski}$^{1,2}$ \\
    \\
    $^{1}$University of Toronto \quad $^{2}$Vector Institute \quad $^{3}$University of British Columbia \quad $^{4}$Google Deepmind
}

\begin{document}
\maketitle

\begin{figure}[H]
    \centering
    \includegraphics[width=\linewidth]{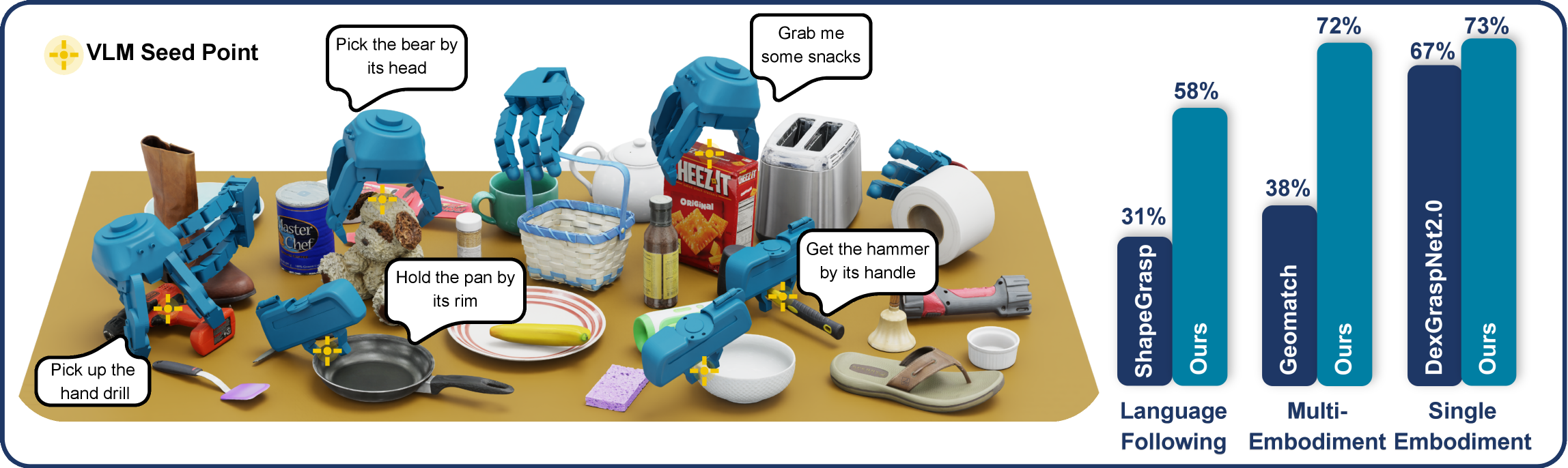}
    \caption{Example grasps and performance summary of our proposed method, \METHOD.}
    \label{fig:main_figure}
\end{figure}

\begin{abstract}
Practical robotic grasping in complex scenes requires both 3D spatial reasoning and alignment with task-specific requirements. Vision-language models (VLMs) offer a natural way to specify these requirements using language, but existing approaches either use a VLM to predict the grasp directly with limited spatial awareness, or train the VLM together with the grasping model, which requires significantly more data and compute. 
These limitations impede performance and have prevented scaling to multiple embodiments in complex scenes.
We address this by proposing SeededGrasp, a novel data-efficient framework that enables a VLM to predict a seed point to be used as conditioning for a subsequent lightweight grasp-generation model.
Our architecture decouples high-level semantic reasoning from low-level geometric execution, enabling multi-embodiment support while bypassing the need for expensive end-to-end training.
To enable training such models, we release the first multi-embodiment tabletop grasping dataset comprising over 2.5M grasps in cluttered scenes. 
Experimental results demonstrate that our approach outperforms existing baselines, achieving 72\% success in simulation and 78\% in real-world grasping experiments.
See our project site for data and code: \href{https://uoft-isl.github.io/seeded-grasp/}{https://uoft-isl.github.io/seeded-grasp/}.
\end{abstract}

\keywords{Dexterous Grasping, Natural Language, Multi-Embodiment}


\section{Introduction}
\vspace{-5pt}
Dexterous grasping is a fundamental requirement for general manipulation but remains a challenging task. To act effectively in the physical world, a robot must make contact with objects in a manner that is precise, stable, and task-aware. Additionally, in contact-rich, multi-object environments such as cluttered scenes, models must respect the objects' surroundings~\cite{sundermeyer2021contactgraspnet,zhang2024dexgraspnet}. 
Moreover, because downstream tasks impose different functional requirements, the ability to follow language instructions becomes highly advantageous~\cite{li2024shapegrasp,tang2023graspgpt,he2025dexvlg}.
This challenge is further compounded by the wide variety of robotic end-effectors. 
Simple two-finger grippers, while widely used, struggle with more complex object geometries because they are limited to pinch grasps. In contrast, grippers with high dexterity, such as the Allegro Hand, can approach objects in multiple ways but are difficult to control~\cite{chen2025dexonomy,feix2016grasp}.
Supporting multiple embodiments offers the additional advantage of improving grasping performance, making a single multi-embodiment model more useful and data-efficient than specialized ones~\cite{fei2025trograsp,li2023gendexgrasp,geomatch,wei2024drograsp}.
Together, these considerations motivate the problem addressed in this paper: designing a system that connects high-level linguistic intent to low-level contact geometry, while remaining effective in cluttered scenes and across diverse embodiments. 

Existing approaches address only parts of this problem, and important gaps remain. Some methods focus on isolated objects, which simplifies perception and contact reasoning but does not capture the complexity of real-world cluttered scenes~\cite{wang2022dexgraspnet,mousavian20196dofgraspnet}. Works that study language-grounded grasping systems often require large-scale annotated datasets and expensive end-to-end training involving a VLM~\cite{he2025dexvlg,deng2025graspvla}, or leverage pre-trained VLMs to predict high-level representations with complex pose optimization pipelines~\cite{li2024shapegrasp,jian2025zerodexgrasp,nguyen2025graspmas}. These approaches do not scale well to cluttered multi-object settings with dense occlusions and/or multi-embodiment grasping where gripper geometries differ greatly and large-scale datasets cannot be collected for every gripper. In addition, existing grasping datasets do not simultaneously provide both cluttered multi-object scenes and multi-embodiment poses~\cite{zhang2024dexgraspnet,wang2022dexgraspnet,casas2024multigrippergrasp}. This mismatch motivates our approach and the creation of a new dataset. 


In this work, we present \METHOD, a modular framework for language-grounded, multi-embodiment grasping in cluttered scenes. Our key idea is to use a VLM to select a task-specific point from the scene point cloud, that indicates where the grasp should be centered. This seed point allows the VLM to isolate the target object and its relevant part while giving the grasping model flexibility in how to grasp it. Conditioned on this seed point, we train a lightweight flow matching model for grasp generation~\cite{flowmatching}. To train our model, we develop a synthetic data generation pipeline and curate a new dataset containing 2.56M grasps across 610 scenes, 334 objects, and 3 grippers. We evaluate \METHOD\space both in simulation, where it outperforms all baselines, and real-world experiments, where we demonstrate its sim-to-real capabilities. 

In summary, our contributions are twofold: 
\begin{enumerate}[label=\arabic*), leftmargin=*, topsep=0pt, itemsep=0pt] 
    \item A novel approach using a VLM to zero-shot predict a seed point which conditions a lightweight flow-matching model for efficient, multi-embodiment grasp generation. 
    \item A new grasping dataset of 2.56M grasps across 610 cluttered scenes spanning three grippers. 
\end{enumerate}

\begin{figure}
    \centering
    \includegraphics[width=\linewidth]{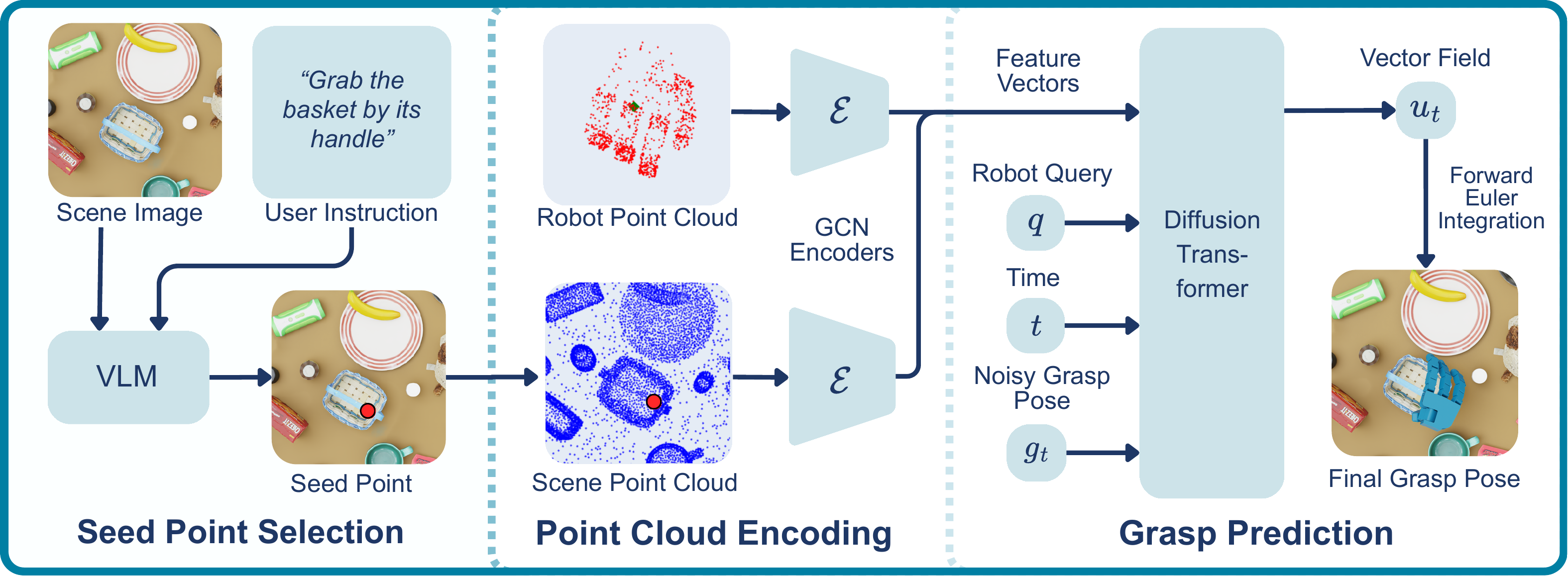}
    \caption{We use a VLM to select a seed point. The point cloud encoding and grasp prediction sections are trained in an end-to-end fashion and predict a grasp pose conditioned on the seed point. During training we use seed points derived from the dataset.}
    \vspace{-15pt}
    \label{fig:model_arch}
\end{figure}

\section{Related Work}
\label{sec:related_work}
\vspace{-5pt}
\paragraph{Analytic and simulation-based grasp synthesis.}
Early work on robotic grasping commonly formulates grasp generation as a search or optimization problem with analytic stability criteria~\cite{miller2004graspit}. Recent optimization-based approaches have substantially improved the speed and quality of dexterous grasp synthesis~\cite{liu2021dfc,turpin2023fastgraspd,chen2024bodex}. While these methods are still relatively unwieldy at inference time and generally do not work on cluttered scenes or incorporate semantic intent from language, they are very valuable for generating datasets of valid grasps. 

\paragraph{Datasets.}
Progress in grasp learning has been closely tied to dataset quality. Large synthetic datasets have enabled major advances in both parallel-jaw and dexterous grasping~\cite{wang2022dexgraspnet,mahler2017dexnet2,eppner2021acronym}, and more recent datasets have expanded to cluttered scenes~\cite{zhang2024dexgraspnet,fang2020graspnet1billion} and multiple embodiments~\cite{li2023gendexgrasp,casas2024multigrippergrasp}. However, none of the existing datasets provide data for multi-embodiment grasping in cluttered scenes.
In this work, we build our dataset upon MultiGripperGrasp~(MGG), which itself filters grasps generated by GraspIt~\cite{miller2004graspit} in simulation for stability across various objects and grippers.  

\paragraph{Grasping in cluttered scenes.}
A major challenge in manipulation is moving from isolated objects in mid-air to cluttered, contact-rich scenes where the model must reason about occlusion and interactions with surrounding objects. Recent learning-based methods address this by predicting grasp distributions directly from scene observations, often using point clouds or depth images as input~\cite{sundermeyer2021contactgraspnet}. For dexterous hands, this has led to generative models, contact-aware representations, and collision-aware formulations that better capture the complexity of multi-object scenes~\cite{zhang2024dexgraspnet,lundell2021ddgc,li2022hgcnet}. Even so, most of this work focuses on geometric feasibility in clutter and is typically developed for a fixed gripper embodiment whereas our work trains a shared model across multiple embodiments.

\paragraph{Multi-embodiment grasping.}
Another research direction studies how grasping models can generalize across different grippers and dexterous hands. Rather than training a separate policy for each end-effector, these methods condition prediction on robot geometry or kinematic structure~\cite{fei2025trograsp,wei2024drograsp,shao2020unigrasp,xu2021adagrasp}. Some approaches use hand-agnostic intermediate representations, such as contact maps or keypoints, and then solve for specific gripper configurations~\cite{li2023gendexgrasp,geomatch}. Others directly encode morphology and learn grasp generation end-to-end~\cite{wei2024drograsp,wei2024geomatchpp,freiberg2024diffusionmultiembodiment}. Collectively, these methods show that conditioning on gripper structure and multi-embodiment training can improve grasp quality. However, they are primarily focused on generating geometrically valid grasps and do not address multi-object scenes and language grounding like our work.

\paragraph{Language-grounded grasping.}
Natural language is an intuitive interface for specifying what object to grasp and how. Supporting such conditioning has motivated language-grounded methods. Several recent approaches~\cite{li2024shapegrasp,nguyen2025graspmas} use VLMs as a plug-and-play module to infer task-relevant regions, affordances, or object parts from text and subsequently rely on heuristics to propose and finetune grasps. These often lead to imprecise results. Other works build larger systems that combine VLM-based semantic grounding with generative modeling techniques like flow matching~\cite{flowmatching}, training the entire pipeline end-to-end~\cite{he2025dexvlg,deng2025graspvla,black2024pi0,wen2024tinyvla}. However, these systems require large-scale, language annotated datasets and are expensive to train. Our method bridges both approaches by introducing a seed point as an efficient interface connecting a pretrained VLM with a learned flow matching grasp generation module, eliminating the need for a language-annotated dataset.


\section{Dataset Generation}
\label{sec:data_gen}
\vspace{-5pt}

To train our seed point conditioned grasp prediction model, we require a dataset of valid grasps in cluttered scenes for each of the grippers. 
Rather than generating raw grasp poses from scratch, we build upon the MultiGripperGrasp (MGG) dataset~\cite{casas2024multigrippergrasp}, which provides millions of synthetic, mid-air grasp poses across a diverse set of grippers and objects. Adapting this dataset involves generating cluttered scenes and transforming these stable mid-air poses into valid in-scene grasps for training.

The MGG dataset was re-filtered using its original infrastructure, which includes mid-air gravity and 6-axis grasp tests~\cite{casas2024multigrippergrasp}, and several object and gripper models were refined to ensure high grasp quality. Following this filtering process, the three grippers yielding the highest volume of quality data while representing a good mix of both parallel-jaw and dexterous grasping kinematics were selected: the Franka Panda, Robotiq 3-Finger, and Allegro grippers.

Cluttered scenes were generated using a split of 284 objects for training and 50 unseen objects for evaluation from the Google Scanned Objects~\cite{GSO} and YCB~\cite{calli2015ycb} datasets. These objects encompass both simple shapes and irregular geometries. Each scene was constructed by randomly sampling one to ten objects, initializing them in mid-air with random orientations, and dropping them on a tabletop. This procedure yields cluttered scenes featuring both sparse and dense regions. In total, 504 training and 106 evaluation scenes were generated. 

Grasps for each of these scenes were generated by transforming existing MGG grasps and checking if they were viable. For every object present, its associated mid-air grasps were transformed into the scene's global frame. Naturally, many of them are infeasible in the cluttered environment due to collisions with the table or other objects. Because conducting dynamic, in-scene lift tests for millions of grasps is computationally prohibitive, a set of heuristic collision checks efficiently determined feasibility. Specifically, the gripper had to approach the object from a viable angle ($\geq0.5\text{ rad}$), all gripper links had to maintain clearance above the table ($\geq 0.5 \text{ cm}$), and the gripper should not excessively intersect with objects (by enforcing a $10 \text{ mm}$ interpenetration threshold). 

For each scene, we generate a 2048-point point cloud by fusing four RGB-D images captured at cardinal directions and downsampling it using farthest point sampling with a bias away from the table plane. We capture a bird's-eye view~(BEV) for seed point selection by the VLM during inference for this paper, but our approach generalizes to arbitrary viewpoints where target objects are visible.


\begin{table}[h]
    \centering
    \renewcommand{\arraystretch}{1.2} 
    
    \begin{tabular}{l||c|c|c|c}
    \hline
    \textbf{Robot Model} & \textbf{Total Grasps} & \textbf{After Refiltering} & \textbf{In-Scene (Train)} & \textbf{In-Scene (Eval)} \\ \hline \hline
    \rowcolor{uoftsecondaryblue_85_tint} 
    Franka Panda      & 2.76 M & 1.57 M  & 1.61 M & 469 K \\
    Robotiq 3 Finger  & 2.72 M & 990 K  & 246 K  & 27 K  \\
    \rowcolor{uoftsecondaryblue_85_tint} 
    Allegro           & 2.77 M & 766 K  & 161 K  & 48 K  \\ \hline
    \end{tabular}
    \caption{Grasp Dataset Statistics by Robot Model. While all three grippers started with the same amount of grasps, Robotiq and Allegro grasps were viable in scene at a significantly lower frequency.}
    \vspace{-10pt}
    \label{tab:data_stats}
\end{table}

In total, our dataset consists of 2.56 million grasp poses split between training and evaluation sets for the three robot grippers~(see Table \ref{tab:data_stats}). Notably, the dataset distribution is skewed toward the Franka Panda gripper. Its simpler kinematics allow for a higher percentage of grasp poses to pass filtering. While our pipeline allows us to balance the distribution with further targeted data generation, further generation was found to be unnecessary for the scope of this work (\Cref{sec:results_ablations}).
	

\section{Model}
\label{sec:model}
\vspace{-5pt}

As shown in Figure \ref{fig:model_arch}, our grasp generation pipeline consists of two stages: the first predicts the seed point, and the second predicts the grasp pose based on the seed point prediction. 
This approach is inspired by DexGraspNet2.0 (DGN2.0)~\cite{zhang2024dexgraspnet}, but does not require training a seed point prediction model; instead, we use an off-the-shelf VLM to predict a suitable grasping location based on a BEV scene image and user instructions (\Cref{sec:model_infer}). This predicted seed point is then passed to a trained flow-matching model to synthesize a grasp pose for the selected gripper (Sections \ref{sec:model_arch} and \ref{sec:model_train}). This design enables precise natural-language control over grasp generation, incurs low training costs, and does not require generating and filtering numerous candidate poses. Furthermore, the approach is not specific to the selected VLM in this modular design, enabling natural performance scaling from future VLM improvements.  

\subsection{Model Architecture}
\label{sec:model_arch}

\textbf{Point Cloud Encoding.} Both gripper and scene geometries are colourless point clouds, denoted as $p_r \in \mathbb{R}^{1024 \times 3}$ and $p_c \in \mathbb{R}^{2048 \times 3}$ respectively. The point coordinates are Fourier-encoded~\cite{nerf} and concatenated with both the local normal vector ($n \in \mathbb{R}^3$) and the curvature ($\sigma(p) \in \mathbb{R}$), which are estimated via a Principal Component Analysis of local neighborhoods to enrich surface representation. Furthermore, each point's representation in the scene point cloud also contains its distance to the seed point. Following GeoMatch~\cite{geomatch}, Graph Convolutional Networks encode the point clouds using the sixteen nearest neighbors, yielding per-point feature vectors $f_p \in \mathbb{R}^{512}$. To extract global representations for each point cloud, per-point feature vectors are aggregated using both max and mean pooling. Additionally, to capture the local geometry surrounding the target grasp location, features from the sixteen nearest neighbors to the seed point in the scene point cloud are concatenated with the global scene features. A learnable query $q_r \in \mathbb{R}^{512}$ is introduced for each robot geometry to encode the specific gripper selection. Concatenating these supplementary vectors produces the final feature vectors $f_r \in \mathbb{R}^{3 \times 512}$ and $f_c \in \mathbb{R}^{18 \times 512}$ for the robot and scene point clouds respectively.

\textbf{Grasp Pose Parameterization.} Each pose is parameterized as a vector $g = (T, R, \theta)$, where $T \in \mathbb{R}^3$ is the translation, $R \in \text{SO}(3)$ is the rotation, parameterized with Euler angles~\cite{bose2024se3}, and $\theta$ is the joint configuration of the gripper within the world frame. Because the number of actuated joints varies across the evaluated grippers, the pose representation is padded to accommodate the maximum degrees of freedom, which is sixteen joints for the Allegro hand.

\textbf{Grasp Pose Generation.} The distribution of stable grasp poses conditioned on robot and scene features, $p(g \mid f_r, f_c)$, is inherently complex and multi-modal. Consequently, we use flow matching to model and sample from $p(g \mid f_r,f_c)$~\cite{flowmatching}. The flow matching model $u_\theta(t, g_t, f_r, f_c)$ is trained to predict a ground truth vector field $u_t(t, g_t, f_r, f_c)$. Given a time step $t \in [0, 1]$ and conditioning features $f_r$ and $f_c$, the model transforms a noisy grasp pose sample $g_t$ into a sample from the target data distribution. By numerically solving the corresponding Ordinary Differential Equation (ODE) (detailed in \Cref{sec:model_infer}), an initial noise sample $g_0$ is iteratively transformed into a clean grasp pose $g_1$. The underlying network employs a standard Diffusion Transformer (DiT) architecture, which utilizes cross-attention mechanisms to inject the robot and scene conditioning signals~\cite{Peebles2022DiT}.

\subsection{Model Training}
\label{sec:model_train}

\textbf{Data Preparation.}  \METHOD\space is trained on our entire synthesized grasp dataset. 
To train the grasp prediction model without querying a VLM each time, we sample one of the eight points on the target object closest to the gripper's center of palm position to act as a ground-truth seed point. 
We mitigate overfitting to simpler object and gripper geometries, which naturally pass the heuristic filtering at higher rates due to fewer collisions between them, by sampling grasp poses uniformly across grippers, scenes, and objects. To promote generalization, random Z-axis rotations, XY-plane translations, and Gaussian noise are applied to the point clouds and grasp poses where applicable. Furthermore, all grasp poses are min-max normalized to a range of $[-1, 1]$ with distinct bounds for the joint angles of each specific gripper to account for variations in joint limits, scales, and units, facilitating multi-embodiment learning.

\textbf{Training Details.} The network is optimized using a conditional flow-matching loss, $\mathcal{L}_{\text{cfm}}$, computed across the translation, rotation, and joint angle components of the predicted vector field $u_\theta$. The total loss is formulated as a linear combination: $\mathcal{L} = \lambda_{\text{trans}}\mathcal{L}_{\text{trans}} + \lambda_{\text{rot}}\mathcal{L}_{\text{rot}} + \lambda_{\text{joints}}\mathcal{L}_{\text{joints}}$, where the $\lambda=(0.4, 0.04, 0.8)$ coefficients serve as scaling factors to balance the respective terms. Unused joint dimensions for specific grippers are masked during the computation of $\mathcal{L}_{\text{joints}}$, and a subsequent averaging operation ensures that joint losses remain balanced across the different embodiments. Instead of the standard L2 loss, an L1 loss is employed for $\mathcal{L}_{\text{cfm}}$:
\begin{align*}
   \mathcal{L}_{\text{cfm}} = \mathbb{E}_{t \sim \beta(1.5, 1.0), \; (g_t, f_r, f_c) \sim \mathcal{U}} \left[ \left| u_\theta(t, g_t, f_r, f_c) - u_t(t, g_t, f_r, f_c) \right| \right] 
\end{align*}
This substitution improves the precision with which generated grasp poses adhere to scene geometries. Furthermore, the time step $t$ is sampled from a Beta distribution during training to bias the loss toward later time steps ($t \to 1$), where highly accurate vector fields are critical for finalizing precise grasp poses~\cite{black2024pi0}. Classifier-Free Guidance is also integrated to strengthen the model's adherence to the conditioning signals~\cite{ho2022classifierfreediffusionguidance}. By randomly dropping out the conditioning features $f_r$ and $f_c$ during training, the network simultaneously learns both conditional and unconditional vector fields. During inference, these fields are extrapolated to push the generated samples toward stronger conditioning (using $w=1.1$), thereby improving alignment with the surrounding scene geometry.

\subsection{Model Inference}
\label{sec:model_infer}

\textbf{VLM Seed Point Prediction.} During inference, a VLM is prompted to zero-shot identify a stable grasping location that aligns with the user's directive~(prompt detailed in \Cref{sec:vlm_prompt}). The predicted point in the image is subsequently projected onto the 3D scene point cloud and passed to the generation module. VLMs accomplish this task effectively without requiring fine-tuning or few-shot examples. We ablate different models in \Cref{sec:results_ablations} and find that Gemini models perform best.

\textbf{Generating Grasp Poses.} The final grasp pose is generated by denoising an initial pose state conditioned on the chosen seed point, the current prediction point clouds, and the target robot query vector. To simplify the inference trajectory and minimize numerical error accumulation, the initial state $g_0$ is initialized as a fixed canonical grasp pose adjacent to the seed point, rather than pure noise. Forward Euler Integration with a step size of $0.2$ is used to numerically integrate $g_0$ through the predicted vector field $u_\theta(t, g_t, f_r, f_c)$ from $t=0$ to $t=1$. 


\section{Experimental Results}
\label{sec:results}
\vspace{-5pt}

Our approach supports a broader set of features than prior work, so we cannot do a single one-to-one comparison. Therefore, we evaluate our method along three distinct axes to isolate specific features. First, we show that the model is able to generate successful grasps more effectively using seed points than prior work, and that VLMs can produce better seed points than a specially trained seed point generation module~(\Cref{sec:seed_point_sel_cond_grasp_pred_eval}). Secondly, we show that \METHOD\space can effectively generate grasps across all embodiments~(\Cref{sec:eval_multi_embod}). Thirdly, we demonstrate that seed points are a more effective VLM interface for language-conditioned grasping~(\Cref{sec:eval_lang_cond}). We then ablate gripper encoding, VLM model, and dataset size in \Cref{sec:results_ablations}. Finally, we perform real-world experiments and demonstrate sim-to-real capabilities in \Cref{sec:results_real}.
The flow matching model was trained for 48 hours using two A100 GPUs, and Gemini 3.1 Flash is used for seed point prediction in all evaluations. 

\begin{table}[tbp]
    \centering
    \begin{minipage}[t]{0.45\textwidth}
        \centering
        \renewcommand{\arraystretch}{1.2}
        \begin{tabular}{l||c|c}
        \hline
        \multicolumn{1}{c||}{\multirow{2}{*}{\textbf{Method}}} & \multicolumn{2}{c}{\textbf{Success (\%) (Allegro Data Only)}} \\ \cline{2-3}
        \multicolumn{1}{c||}{} & Graspness Seeds & VLM seeds \\ \hline \hline
        \rowcolor{uoftsecondaryblue_85_tint} DGN2.0 & 53.15 & 66.67 \\ 
        Ours & \textbf{66.22} & \textbf{72.97} \\ \hline
        \end{tabular}
        \label{tab:sim_eval_res_seeds}
    \end{minipage}
    \hfill 
    \begin{minipage}[t]{0.52\textwidth}
        \centering
        \renewcommand{\arraystretch}{1.2}
        \begin{tabular}{l||c|c|c}
        \hline
        \multicolumn{1}{c||}{\multirow{2}{*}{\textbf{Method}}} & \multicolumn{3}{c}{\textbf{Success (\%)}} \\ \cline{2-4}
        \multicolumn{1}{c||}{} & Franka & Robotiq & Allegro \\ \hline \hline
        \rowcolor{uoftsecondaryblue_85_tint} Geomatch & 25.35 & 35.71 & 37.97 \\
        Ours & \textbf{71.38} & \textbf{72.16} & \textbf{72.07} \\ \hline
        \end{tabular}
    \end{minipage}
    \caption{Left: Our method and DGN2.0~\cite{zhang2024dexgraspnet} trained only on Allegro data across seed points from the Graspness module of DGN2.0 and a VLM. Right: On multi-embodiment data, \METHOD\space outperforms Geomatch~\cite{geomatch} on complex objects in cluttered scenes.}
    \vspace{-15pt}
    \label{tab:sim_eval_res_seeds_morph}
\end{table}

\subsection{Simulation Results}
\label{sec:results_baselines}

\numberedparagraph{Seed Point Selection and Conditional Grasp Prediction:}
\label{sec:seed_point_sel_cond_grasp_pred_eval}
We first compare our flow matching model to DGN2.0 as both can accept the same seed points. As DGN2.0 only supports a single gripper, we train both methods only using Allegro data. \METHOD\space performs better than the baseline in single embodiment grasping by 13\%~(\Cref{tab:sim_eval_res_seeds_morph}) when conditioned on seed points generated by the Graspness Score Module from DGN2.0. Further, seed points predicted by Gemini 3.1 Flash show higher success rates for both methods than the specially trained seed generation module in DGN2.0 ~\cite{zhang2024dexgraspnet}~(\Cref{tab:sim_eval_res_seeds_morph}). 
Unlike DGN2.0, our model does not need explicit object segmentation masks or an additional feature vector produced by its graspness module and supports multiple end-effectors.

\numberedparagraph{Multi-Embodiment Performance:} 
\label{sec:eval_multi_embod}
As DGN2.0 only supports a single gripper, we compare our method to Geomatch~\cite{geomatch} for multi-embodiment performance~(\Cref{tab:sim_eval_res_seeds_morph}). Our model is able to handle the more complicated objects and cluttered scenes better than Geomatch, which was designed for mid-air single object grasping, increasing the success rate by roughly 35\%. 

This difference is caused by Geomatch's architecture~\cite{geomatch} which supports simultaneous multi-embodiment training by predicting contact points and then relying on an Inverse Kinematics (IK) optimization to determine the joint angles.
We found this to be a fragile process requiring an accurate initial guess that Geomatch's heuristic frequently failed to provide on complex geometries present in our dataset.
This is exacerbated by the fusion of multi-camera point clouds yielding incomplete, hollow objects without a bottom.
More details can be found in \Cref{fig:geomatch_fails} in the Appendix.

\begin{table}
    \centering
    \renewcommand{\arraystretch}{1.2} 
    \begin{tabular}{l||>{\centering\arraybackslash}p{2.3cm}|>{\centering\arraybackslash}p{2.3cm}|>{\centering\arraybackslash}p{3.5cm}} 
    \hline
    \multicolumn{1}{c||}{\multirow{2}{*}{\textbf{Method}}} & 
    \multicolumn{2}{c|}{\textbf{Obj. \& Part Recog. Succ. (\%)}} & 
    \multirow{2}{*}{\parbox{3.8cm}{\centering \textbf{Overall Grasp Succ. (\%)}}} \\ \cline{2-3}
    \multicolumn{1}{c||}{} & Easy Prompt & Difficult Prompt & \\ \hline \hline
    \rowcolor{uoftsecondaryblue_85_tint} ShapeGrasp w/ Obj.Seg. & 57.5 & 51.6 & 31.1 \\
    GraspMAS & 42.5 & 29.0 & 24.5 \\
    \rowcolor{uoftsecondaryblue_85_tint} Ours & \textbf{89.6} & \textbf{80.6} & \textbf{58.2} \\ \hline
    \end{tabular}
    \caption{Language-conditioned grasp quality. Center: \METHOD\space selects the correct object and its part at a higher rate. Right: It achieves a higher success rate of grasps for prompts, where the language instruction was correctly followed.}
    \vspace{-15pt}
    \label{tab:functional_grasp_results}
\end{table}

\numberedparagraph{Language Conditioned Grasp Prompting Methods:}
\label{sec:eval_lang_cond}
We compare \METHOD\space against ShapeGrasp~\cite{li2024shapegrasp} and GraspMAS~\cite{nguyen2025graspmas}, two alternative zero-shot techniques that allow VLMs to predict grasps, on their ability to follow prompts and grasp objects successfully. Unlike our method, both baselines only support parallel-jaw grippers and rely entirely on BEV images of the scene. ShapeGrasp segments the object in the image, then prompts the VLM to choose one that best aligns with the user instruction. GraspMAS utilizes an agentic closed-loop process composed of a Planner, Observer, and Coder to choose an object and predict grasps. 

To assess the method's ability to follow language instructions and produce good grasps, we created a test dataset of 226 pairs of scene-prompt queries. These were split between 164 \textit{Easy} and 62 \textit{Difficult} prompts, which described the task respectively directly or indirectly. See prompt examples in \Cref{sec:language_conditioned_grasp_test_dataset}. Predictions were manually scored for adherence to user instruction.

As shown in \Cref{tab:functional_grasp_results}, \METHOD\space outperforms both baselines in correctly recognizing the target object, identifying the correct part, and producing a valid grasp. Both baselines struggle as they rely exclusively on a BEV image of the scene sent to VLM and are unable to utilize point cloud information. Our seed point interface eliminates the need for complex prompting architectures and avoids the drawbacks of relying exclusively on a VLM for grasp prediction. For more details on their specific failures, see \Cref{sec:baseline_fails}.

\subsection{Ablation Studies}
\label{sec:results_ablations}

\textbf{Gripper Encoding.} The model encodes gripper information using both robot point cloud and a learnable gripper-specific query vector. As shown in \Cref{tab:abl_robot}, using both is important, particularly for the more kinematically complex Robotiq 3-Finger and Allegro grippers. They allow the network to better capture the specific geometric intricacies of each embodiment and generalize across grippers.

\begin{table}[h]
    \centering
    \renewcommand{\arraystretch}{1.2} 
    \begin{tabular}{l||c|c|c}
    \hline
    \multicolumn{1}{c||}{\multirow{2}{*}{\textbf{Method}}} & \multicolumn{3}{c}{\textbf{Success (\%)}} \\ \cline{2-4}
    \multicolumn{1}{c||}{} & Franka Panda & Robotiq 3-Finger & Allegro \\ \hline \hline
    \rowcolor{uoftsecondaryblue_85_tint} \METHOD\space w/o Query Vector & 72.41 & 69.07 & 59.01 \\
    \METHOD\space w/o Robot PC & \textbf{72.76} & 67.53 & 62.16 \\
    \rowcolor{uoftsecondaryblue_85_tint} \METHOD\space & 71.38 & \textbf{72.16} & \textbf{72.07} \\
    \hline
    \end{tabular}
    \caption{Removing either gripper representation decreases the success rate}
    \vspace{-15pt}
    \label{tab:abl_robot}
\end{table}

\textbf{VLM.} Because the proposed framework is agnostic to the specific VLM, the grasp success rates of several prominent VLM families are evaluated to assess the impact of the backbone architecture. \Cref{tab:abl_vlm} shows that Gemini 3.1 Flash outperforms competing architectures and highlights that choosing the right VLM is important given current VLM capabilities~(see \Cref{tab:abl_vlm_grippers} in Appendix for per gripper results). Additionally, using a higher reasoning level did not help performance.

\begin{table}[h]
    \centering
    \renewcommand{\arraystretch}{1.2}
    \begin{tabular}{l||c|c|c}
    \hline
     & \textbf{GPT 5.4} & \textbf{Qwen 3.5 Flash~\cite{qwen35blog}} & \textbf{Gemini 3.1 Flash} \\ \hline \hline
    \rowcolor{uoftsecondaryblue_85_tint} \textbf{Success (\%)} & 42.61 & 54.42 & \textbf{71.87} \\ \hline
    \end{tabular}
    \caption{Avg. Success Rates across all three grippers (Franka, Robotiq 3-Finger, and Allegro) using various VLMs for seed point prediction. Gemini 3.1 Flash outperforms others by a large margin.}
    \vspace{-15pt}
    \label{tab:abl_vlm}
\end{table}

\textbf{Dataset Scale.} To evaluate the impact of data volume on generation quality, the model was retrained using 25\%, 50\%, and 75\% of the training scenes. As the dataset size increases, the performance improves as well but saturates at around 71.5\%, indicating only marginal improvements if the dataset size was further increased (See the \Cref{fig:abl_data} in Appendix for plot). Reducing the number of training scenes disproportionately degrades the performance of the Robotiq 3-Finger and Allegro grippers. This disparity is likely due to the Franka Panda having a higher density of viable grasp poses per scene, making it more robust to dataset downsampling.

\begin{wrapfigure}{R}{0.4\textwidth}
    \centering
    \vspace{-30pt}
    \includegraphics[width=\linewidth, trim=3cm 3cm 3cm 3cm, clip]{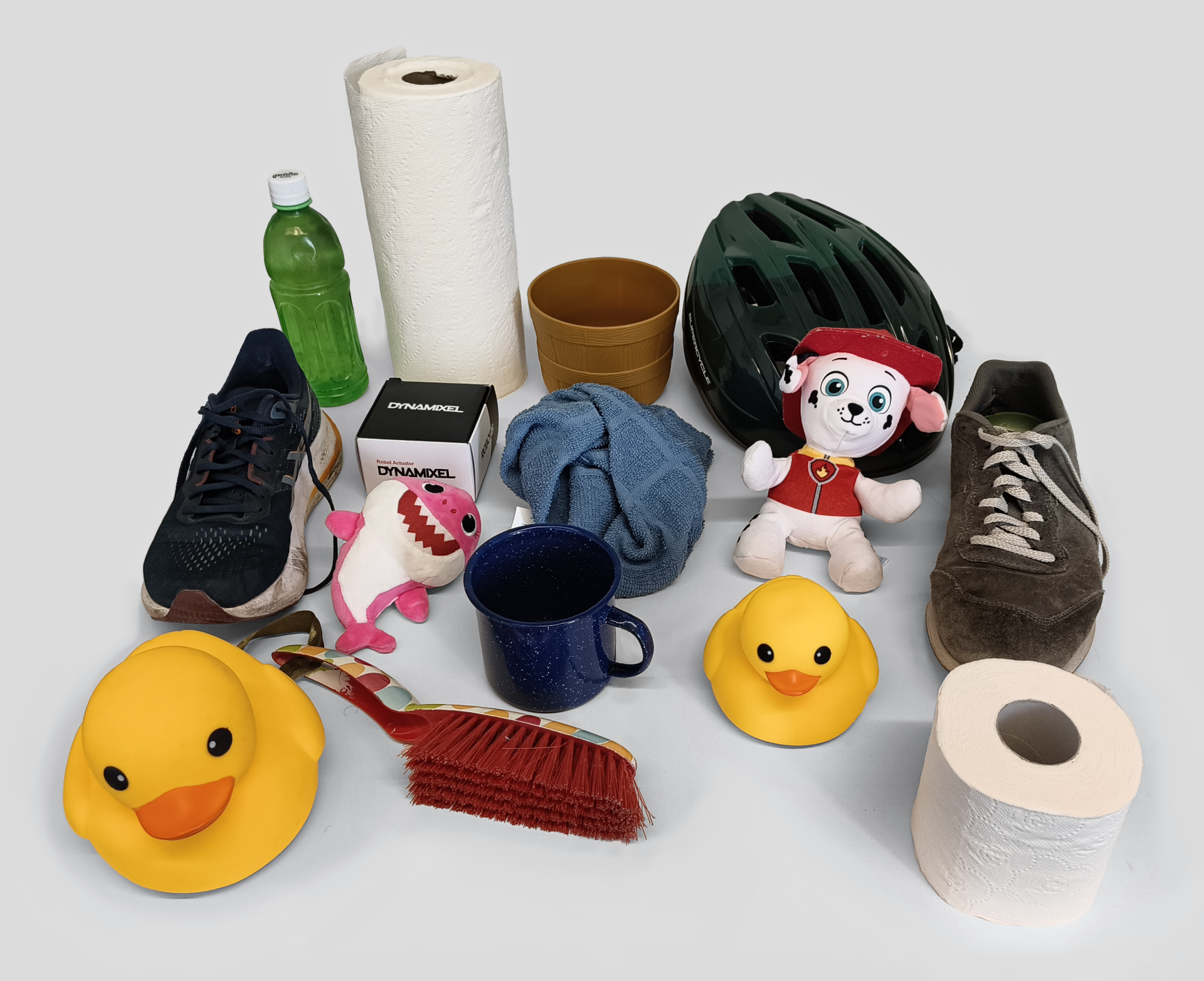}
    \caption{All objects used in real setup}
    \label{fig:real_eval_objects}
    \vspace{-10pt}
\end{wrapfigure}

\subsection{Real-World Evaluation}
\label{sec:results_real}
For the real-world experiment, we evaluate the quality of predicted grasps on a set of real-world objects using a Delto DG-3F-B three-finger gripper.
We used two stationary ZED stereo cameras to generate a point cloud of the table scene.
As the model was not trained on the Delto gripper, we map the Robotiq 3-Finger gripper predictions to the Delto using constant offsets on certain joints and fixed values for the additional joints.
During execution, we approach the objects from above the grasp location and do force-closure after grasping by further closing the two joints at the end of each finger.
We conducted ten trials for each of the fifteen unseen objects with four random distractor objects in the scene.
Relying on the Robotiq 3-Finger predictions has the drawback that the Delto fingers are not as long or thick as the Robotiq ones.
Nevertheless, the model achieved an overall success rate of 78\%. 
See~\Cref{tab:real_performance} in the Appendix for the individual object success rates. 

\section{Limitations}
\label{sec:limitations}
\vspace{-5pt}

In real-world deployments, grippers are mounted on robotic arms that cannot reach every predicted pose, especially when avoiding obstacles. Since our model currently predicts a grasp pose without awareness of the arm's kinematics, it may propose infeasible actions. Future work should address grasp selection and arm motion planning jointly. Additionally, the model’s flexibility could be enhanced by moving beyond single-image VLM conditioning, as incorporating multi-view perspectives could improve spatial reasoning and robustness in cluttered scenes where optimal grasp regions are occluded in some views. Generalization could also be improved by transitioning from embodiment-specific learned queries to a unified representation in joint space, which would facilitate better transfer to unseen hardware across diverse morphologies. Finally, broadening the diversity of the training data is necessary, as the current over-representation of power grasps in the MGG dataset limits the model's effectiveness when handling flat objects that sit flush against a surface.

\section{Conclusion}
\label{sec:conclusion}
\vspace{-5pt}

We present \METHOD, a framework for language-grounded, multi-embodiment grasping in cluttered scenes. \METHOD\space uses a VLM-predicted seed point to encode semantic intent combined with a flow matching model for grasp generation. Trained on our synthetic dataset, it achieves strong grasping performance in both simulation and real-world tests. This demonstrates that seed point conditioning is an effective bridge between language grounding and grasp generation.

\clearpage
\acknowledgments{This research was supported in part by the \href{https://vectorinstitute.ai/}{Vector Institute} and  \href{https://alliancecan.ca/}{Digital Research Alliance of Canada}.}
 

\bibliography{main}  

\newpage
\appendix
\section{Appendix}

\subsection{Evaluation Setup}
Evaluations for all three axes were performed in Isaac Sim with a pass/fail metric for consistency. A grasp is considered a failure if there is significant penetration with objects during initialization, or if the gripper fails to lift the object 30 cm above the table without dropping. These criteria determine the success rates for all baseline and ablation studies.

\subsection{Baseline Setup Details}

\paragraph{DexGraspNet2.0:}To ensure a fair comparison, our model and DGN2.0 were trained and evaluated using models trained on Allegro Hand data, which is morphologically similar to the Leap Hand which was originally used in the baseline by the authors. While the majority of its pipeline operates on the full scene point cloud, it strictly requires explicit object masks to select the initial seed point on the object of interest.

\paragraph{Geomatch:}For Geomatch, which supports simultaneous multi-embodiment training, the method cannot be conditioned with user-generated seed points; instead, it determines its optimal input conditioning autonomously via a top-$k$ parameter. Furthermore, because Geomatch does not support multi-object scenes out-of-the-box, it required object masks at the input and was evaluated exclusively on single-object scenes. And only single object scenes were initialized in the simulation for evaluation as well. 

\paragraph{ShapeGrasp \& GraspMAS:} Both baselines operate with parallel-hand grippers and rely entirely on Bird's Eye View (BEV) images of the scene. ShapeGrasp works only with single object images and was provided with object segmentation masks. It has two modes of operation, 2D and 3D. For our evaluations, ShapeGrasp was evaluated in its 2D mode using \texttt{gemini-3-flash-preview} (identical to the model used in our method), as its native 3D mode failed too often during testing. GraspMAS, which utilizes an agentic closed-loop workflow consisting of a Planner, Observer, and Coder, was evaluated using \texttt{gpt-4o}, as this model yielded the best results for their specific pipeline.

\newpage

\subsection{Language Conditioned Grasp Test Dataset}
\label{sec:language_conditioned_grasp_test_dataset}
As described in \Cref{sec:results_baselines}, we created two sets of prompts to test the performance of the methods at identifying correct object and its part to accomplish the grasping task. 

\paragraph{Easy Prompts:}The 'easy' set of prompts were designed such that it required little to no thinking on the method's part in identifying the correct object or part of the object. The part to be grabbed was simply given in the prompt. Examples of such prompts are given below. 

\begin{figure}[H]
    \centering
    \begin{subfigure}[b]{0.4\linewidth}
        \includegraphics[width=\linewidth]{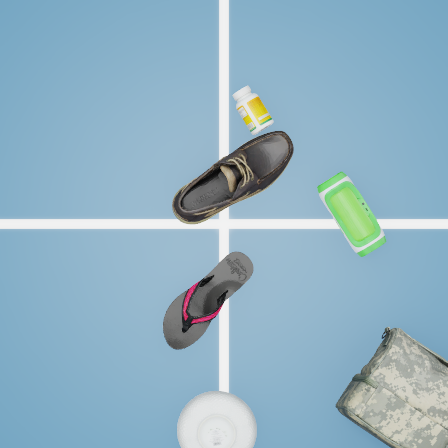}
        \label{fig:easy-1}
        \caption{"Grab the flip-flop by the pink and black straps"}
    \end{subfigure}
    \hfill
    \begin{subfigure}[b]{0.4\linewidth}
        \includegraphics[width=\linewidth]{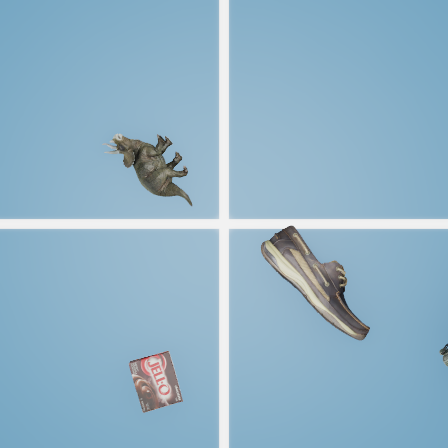}
        \label{fig:easy-2}
        \caption{"Grasp the brown shoe by the front toe box"}
    \end{subfigure}
    \hfill
    \begin{subfigure}[b]{0.4\linewidth}
        \includegraphics[width=\linewidth]{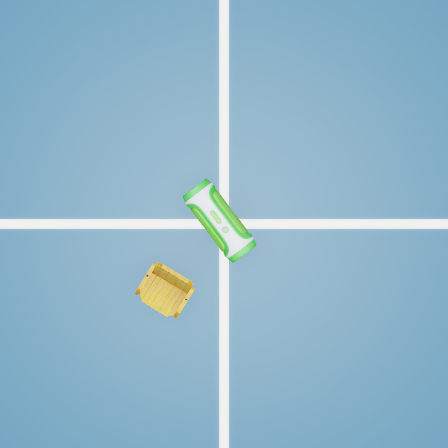}
        \label{fig:easy-3}
        \caption{Grasp the green and white speaker around the center of its cylindrical body}
    \end{subfigure}
    \hfill
    \begin{subfigure}[b]{0.4\linewidth}
        \includegraphics[width=\linewidth]{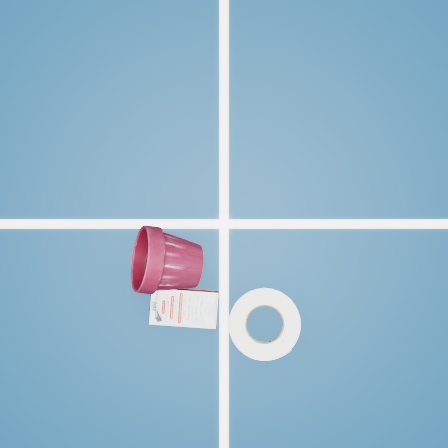}
        \label{fig:easy-4}
        \caption{Grasp the red flower pot by the open top rim on the left}
    \end{subfigure}

    \caption{Examples of easy prompts for objects in scenes}
    \label{fig:easy_prompts_examples}
\end{figure}

\newpage

\paragraph{Difficult Prompts:}The 'difficult' set of prompts were designed such that it requires thinking on the method's part in identifying the correct object or part of the object. The prompt describes a task to be accomplished, the method finds the best part in the object to grab on to. 

\begin{figure}[H]
    \centering
    \begin{subfigure}[b]{0.4\linewidth}
        \includegraphics[width=\linewidth]{assets/easy_example_1.png}
        \label{fig:difficult-1}
        \caption{"Grab anywhere but where the shoe laces are tied"}
    \end{subfigure}
    \hfill
    \begin{subfigure}[b]{0.4\linewidth}
        \includegraphics[width=\linewidth]{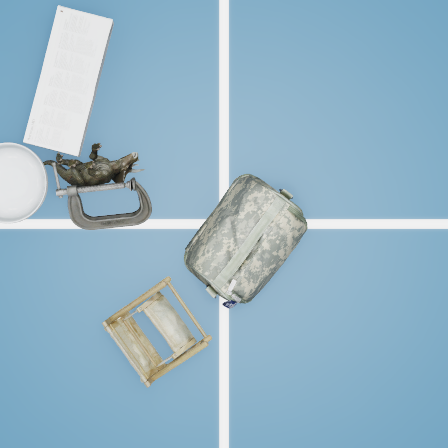}
        \label{fig:difficult-2}
        \caption{"Hold the clamp while I unscrew the handle."}
    \end{subfigure}
    \hfill
    \begin{subfigure}[b]{0.4\linewidth}
        \includegraphics[width=\linewidth]{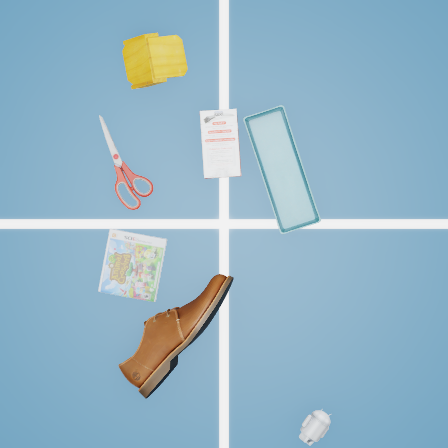}
        \label{fig:difficult-3}
        \caption{Grasp the scissors so you could easily cut something with them.}
    \end{subfigure}
    \hfill
    \begin{subfigure}[b]{0.4\linewidth}
        \includegraphics[width=\linewidth]{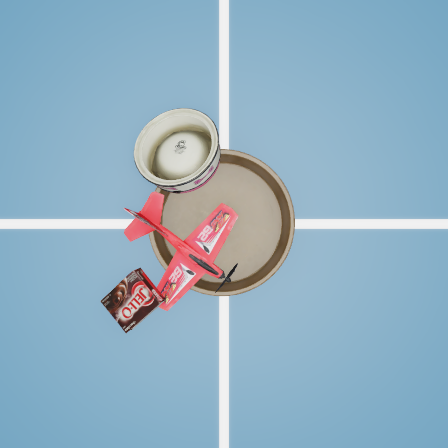}
        \label{fig:difficult-4}
        \caption{Grasp the plane toy where the passenger sits}
    \end{subfigure}

    \caption{Examples of difficult prompts for objects in scenes}
    \label{fig:difficult_prompts_examples}
\end{figure}

For example, in \Cref{fig:difficult_prompts_examples}a, the method should point at any point other than where the laces are. In example \Cref{fig:difficult_prompts_examples}b, the method should grab on to C-shaped body, in example \Cref{fig:difficult_prompts_examples}c, the red handles of the scissors must be grabbed. 

\subsubsection{Failure Cases for Baselines} 
\label{sec:baseline_fails}
\begin{figure}
    \centering
    \begin{subfigure}[b]{0.45\textwidth}
        \centering
        \includegraphics[width=\textwidth, trim=0 0 0 0, clip]{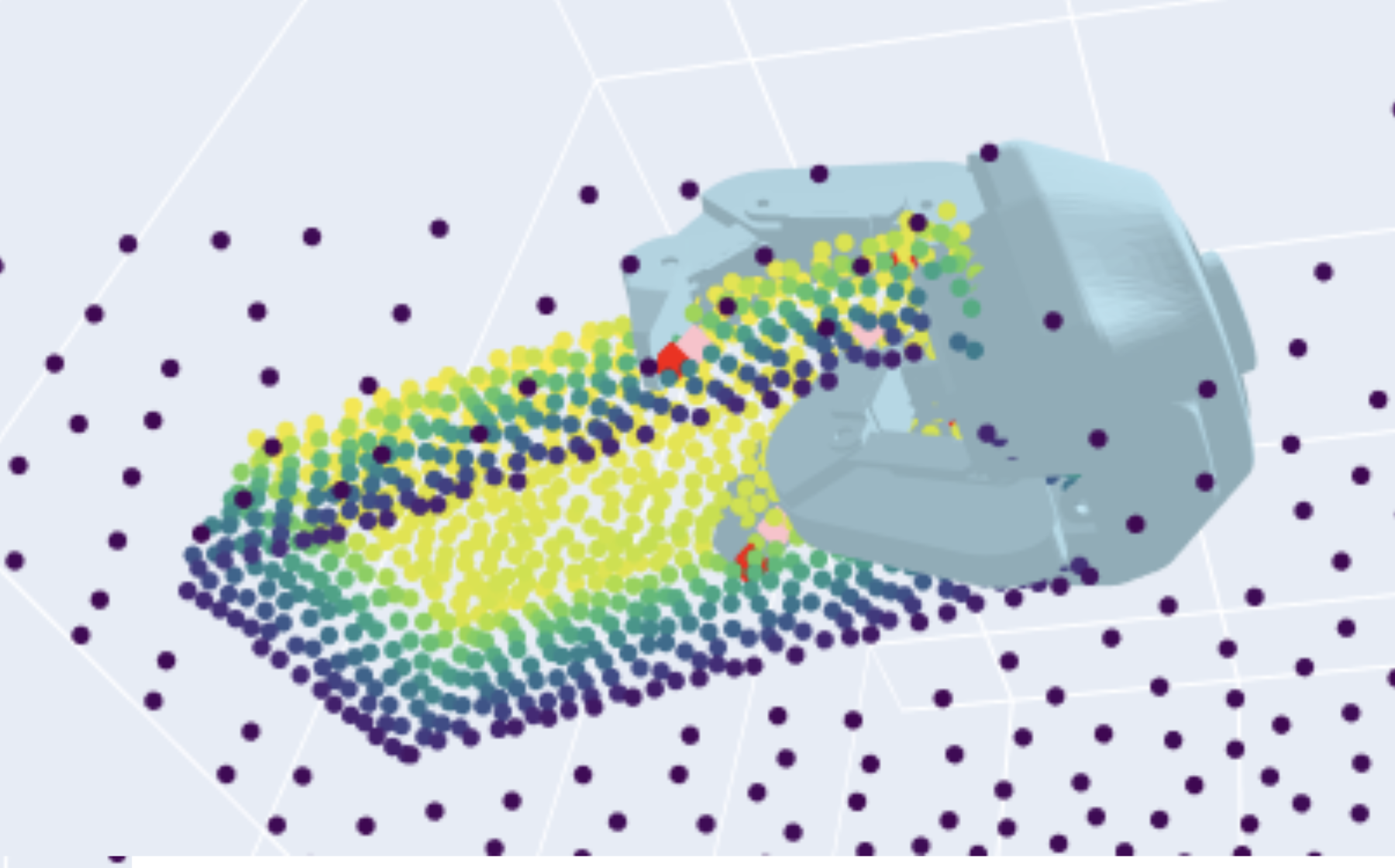}
        \captionsetup{justification=centering}
        \caption{Failure due to partial scene point cloud}
    \end{subfigure}
    \hfill
    \begin{subfigure}[b]{0.4\textwidth}
        \centering
        \includegraphics[width=\textwidth, trim=0 0 0 0, clip]{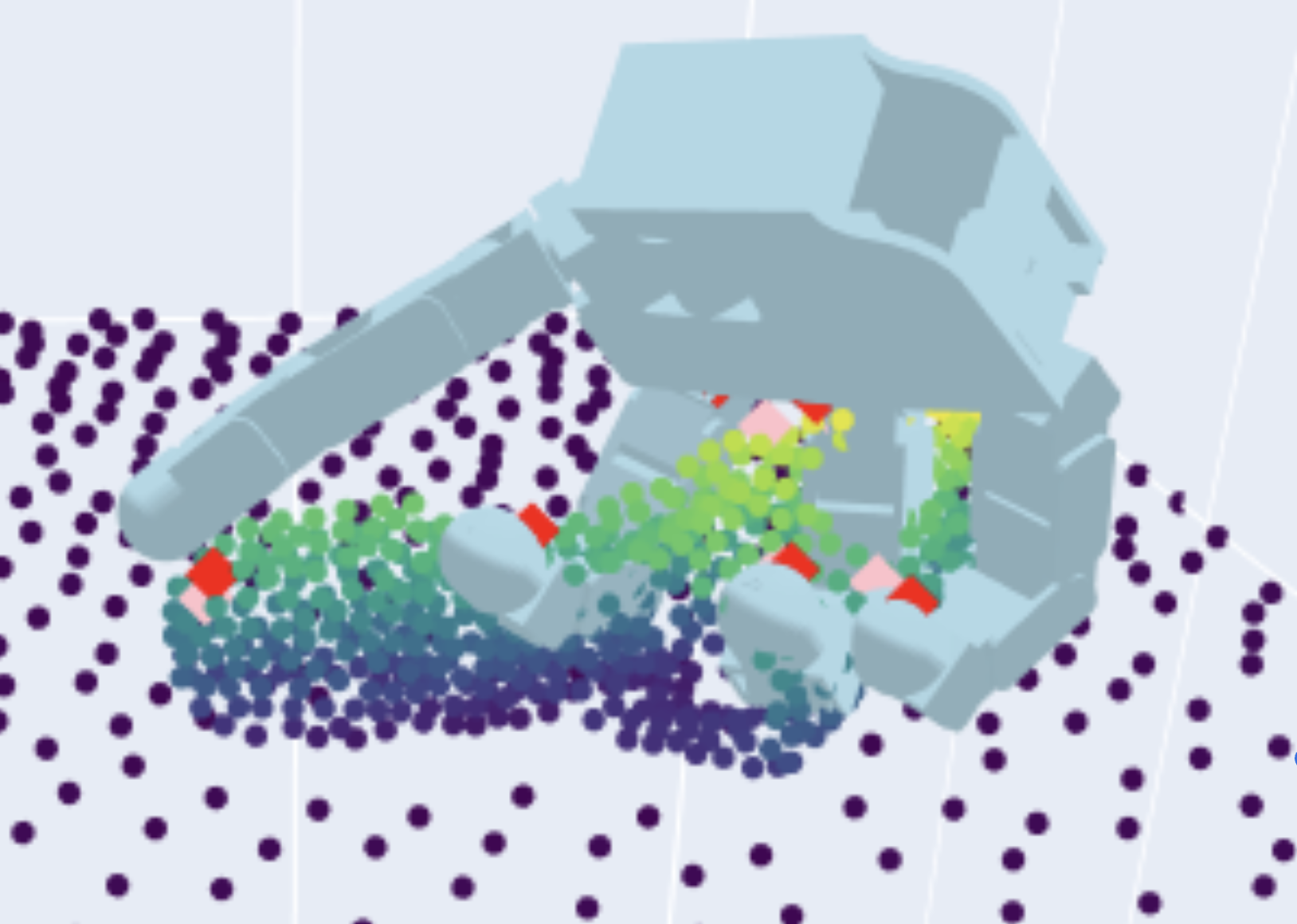}
        \captionsetup{justification=centering}
        \caption{Fail due to complicated object geometry}
    \end{subfigure}
    \caption{Examples of Geomatch failures}
    \label{fig:geomatch_fails}
\end{figure}

\paragraph{Geomatch:} Most of the failures with this method result from object penetrations caused by poor interactions with the IK optimization algorithm in two ways. First, fusing multiple camera views results in  partial point clouds with missing geometry underneath the object sitting on the table. Second, our dataset contains many complicated non-convex objects. Examples for both are provided in \Cref{fig:geomatch_fails}

\begin{figure}
    \centering
    \begin{subfigure}{0.24\linewidth}
        \includegraphics[width=\linewidth, trim=7.5cm 7.5cm 7.5cm 7.5cm, clip]{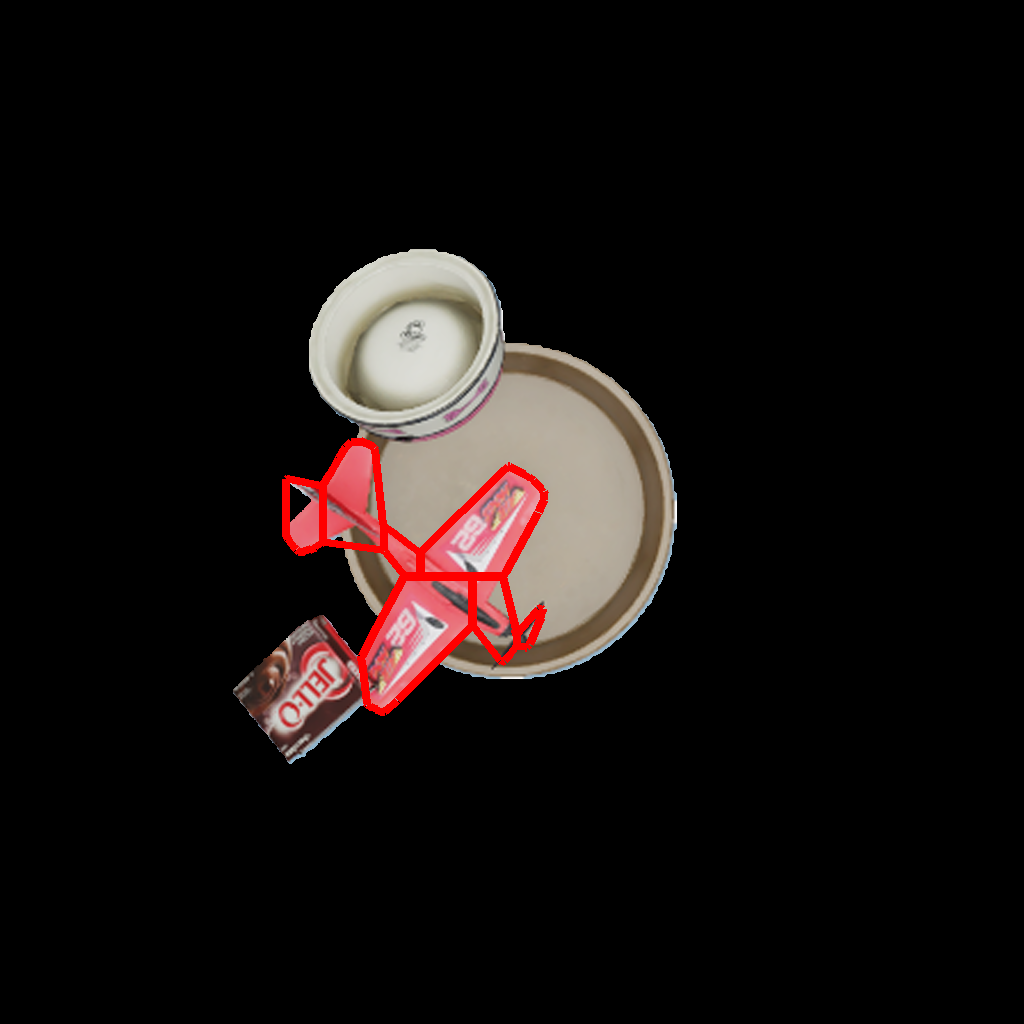}
        \caption*{}
    \end{subfigure}
    \hfill
    \begin{subfigure}{0.24\linewidth}
        \includegraphics[width=\linewidth, trim=7.5cm 7.5cm 7.5cm 7.5cm, clip]{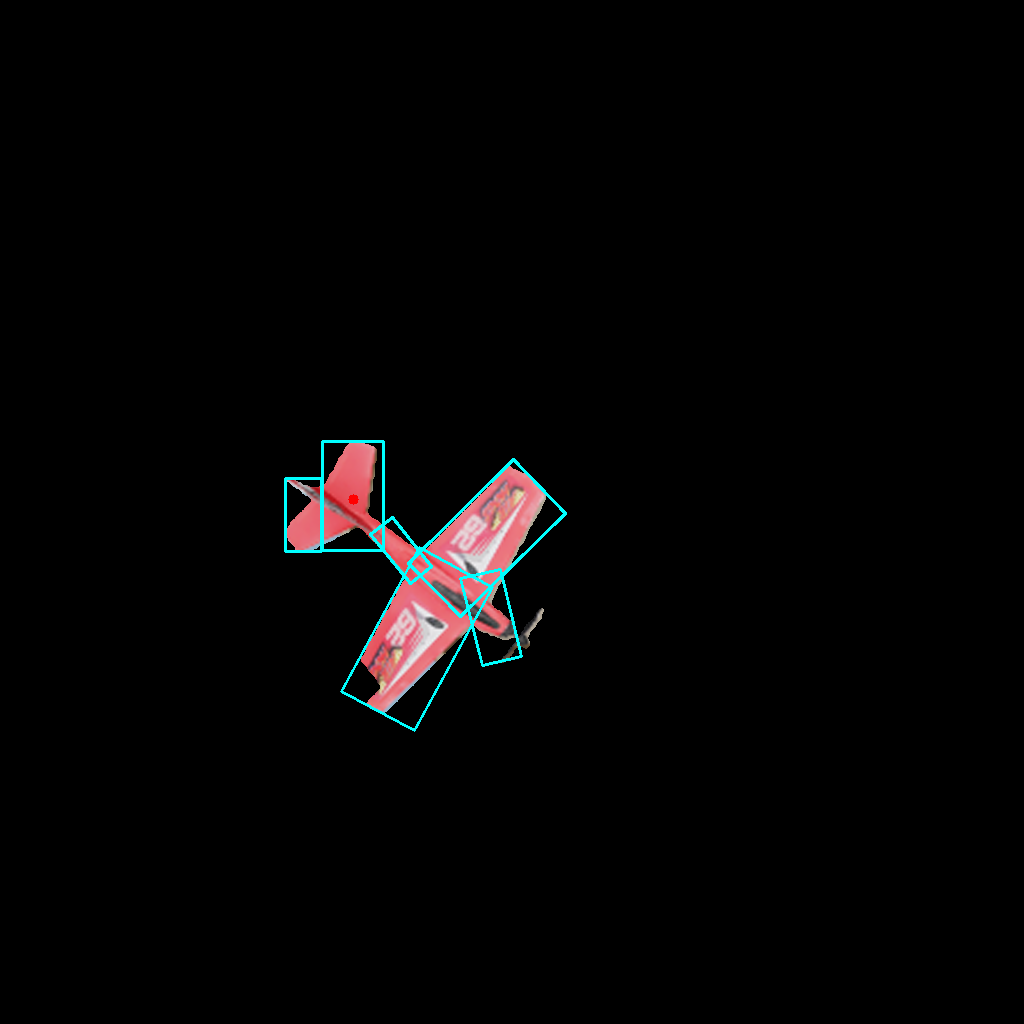}
        \caption*{}
    \end{subfigure}
    \hfill
    \begin{subfigure}{0.24\linewidth}
        \includegraphics[width=\linewidth, trim=2.5cm 10cm 10cm 2.5cm, clip]{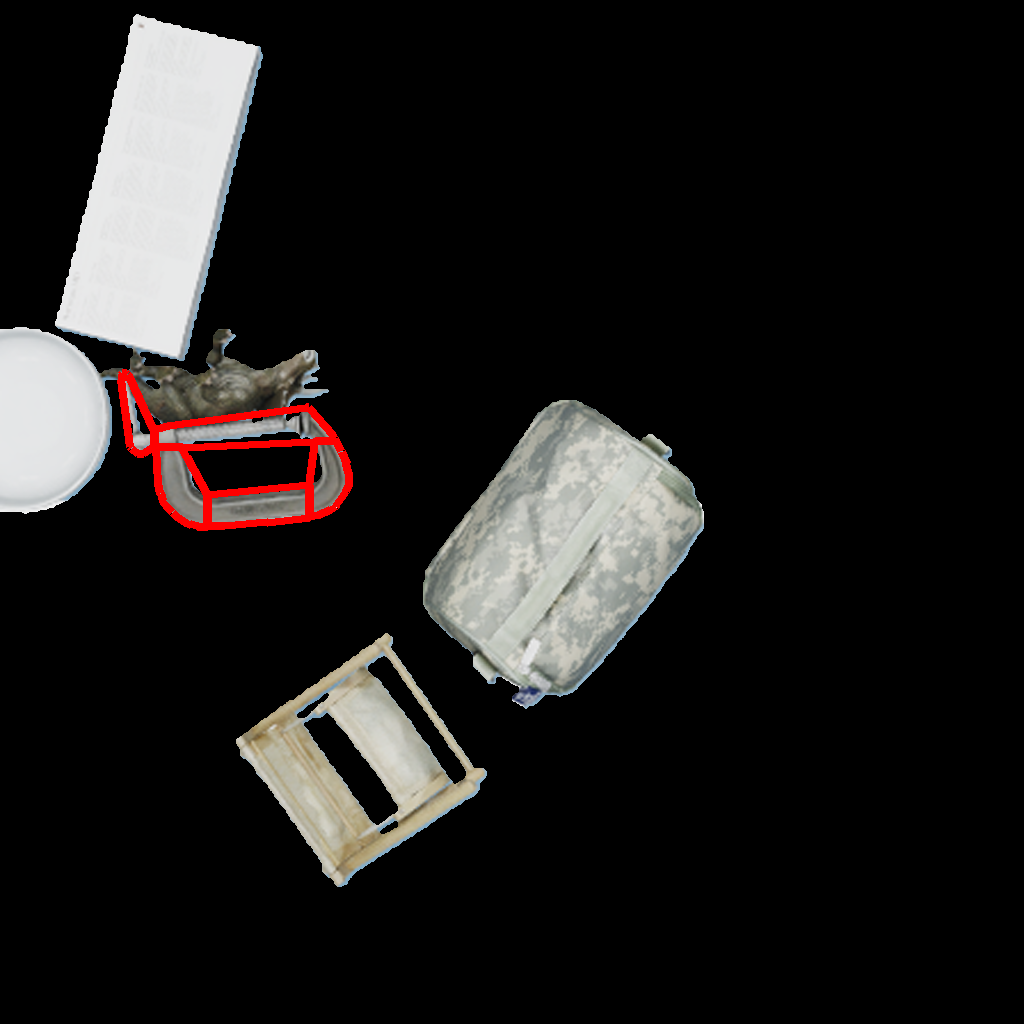}
        \caption*{}
    \end{subfigure}
    \hfill
    \begin{subfigure}{0.24\linewidth}
        \includegraphics[width=\linewidth , trim=2.5cm 10cm 10cm 2.5cm, clip]{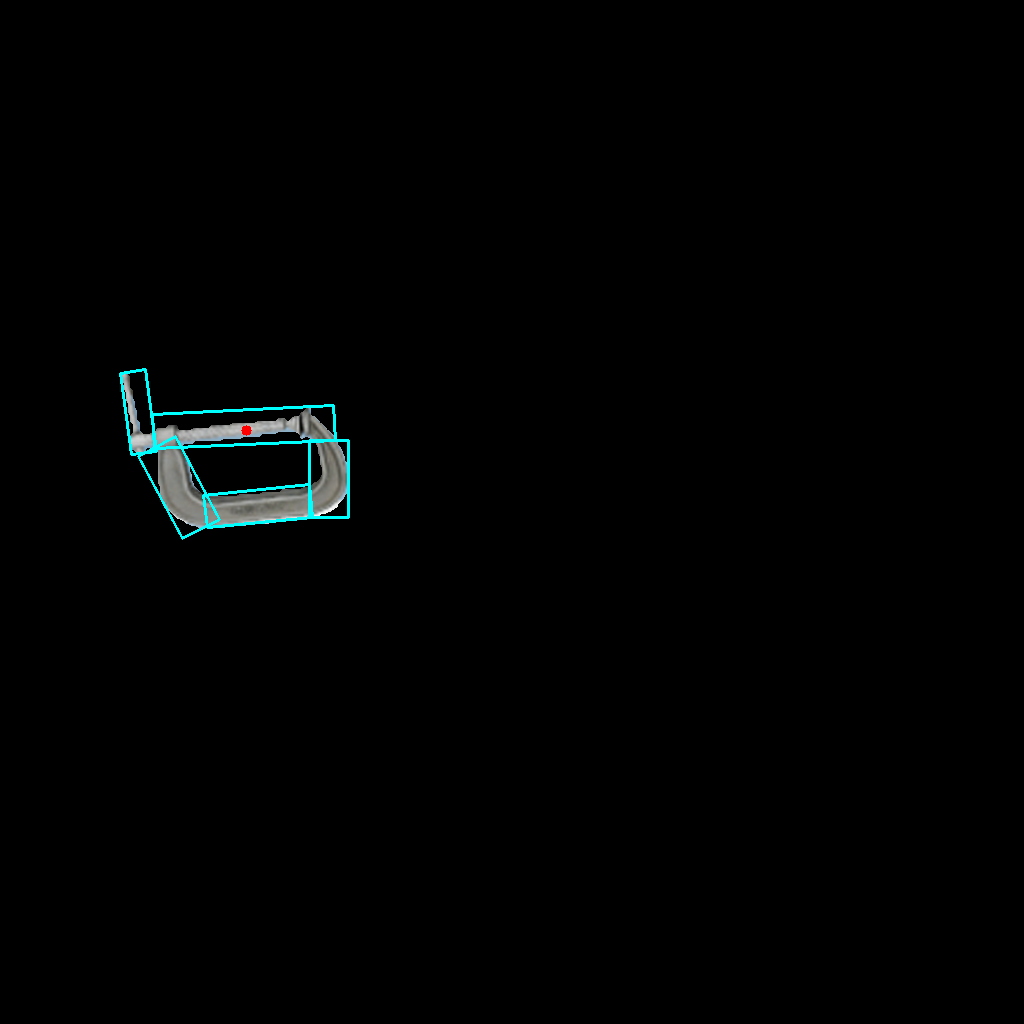}
        \caption*{}
    \end{subfigure}
    \caption{Good examples of part segmentation by Shapegrasp. Objects are split into reasonable number of parts which can be easily identified by their particular purpose}
    \label{fig:shapegrasp_good_examples}
\end{figure}

\begin{figure}
    \centering
    \begin{subfigure}[b]{0.24\textwidth}
        \centering
        \includegraphics[width=\textwidth, trim=5cm 5cm 12.5cm 12.5cm, clip]{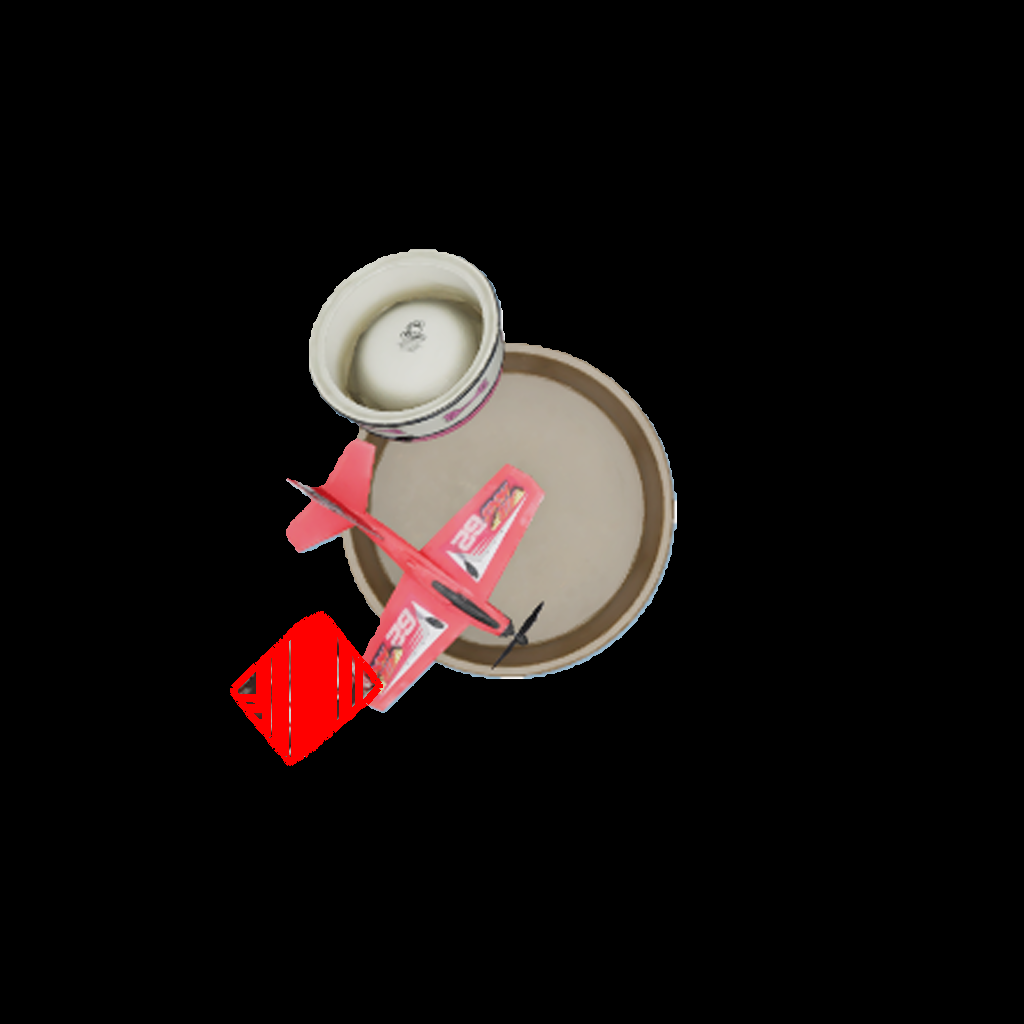}
        \caption*{}
    \end{subfigure}
    \hfill
    \begin{subfigure}[b]{0.24\textwidth}
        \centering
        \includegraphics[width=\textwidth, trim=5cm 5cm 12.5cm 12.5cm, clip]{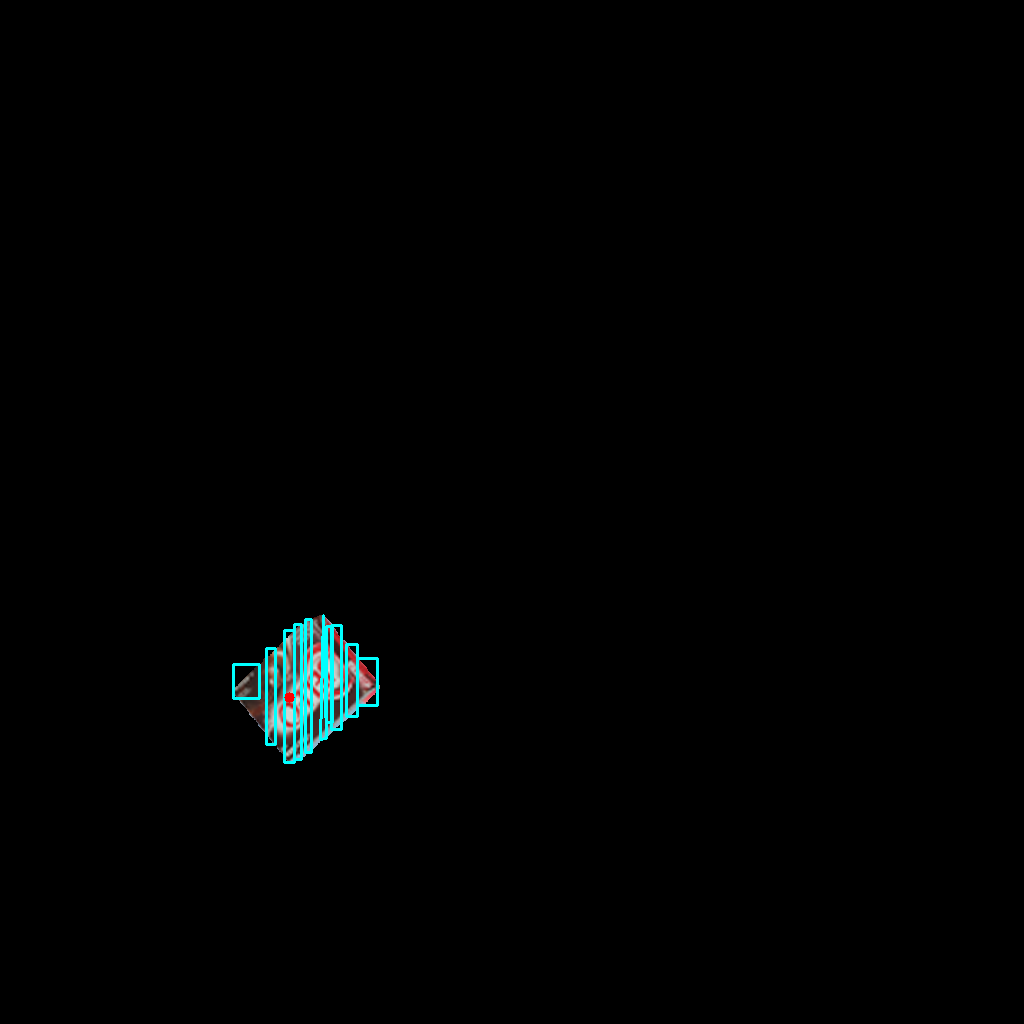}
        \caption*{}
    \end{subfigure}
    \hfill
    \begin{subfigure}[b]{0.24\textwidth}
        \centering
        \includegraphics[width=\textwidth,trim=5cm 5cm 10cm 10cm, clip]{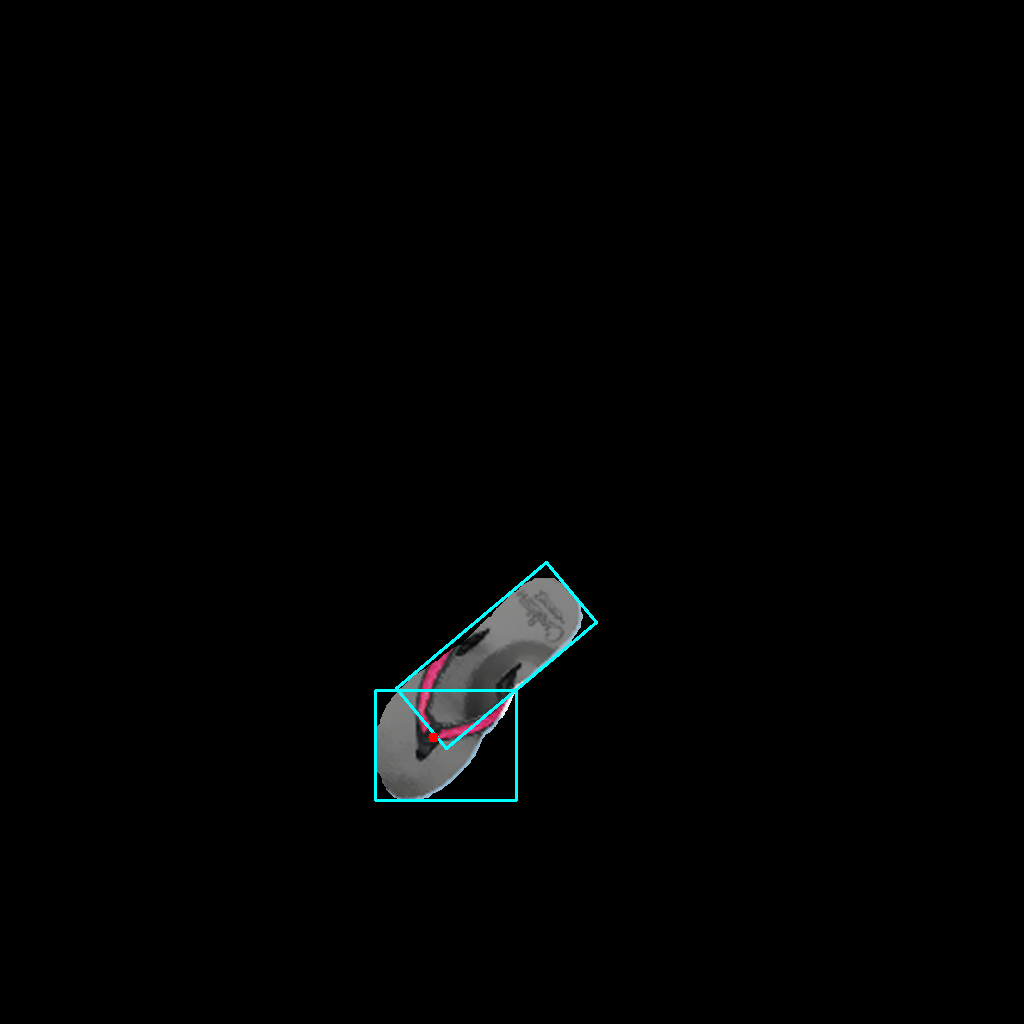}
        \caption*{}
    \end{subfigure}
    \hfill
    \begin{subfigure}[b]{0.24\textwidth}
        \centering
        \includegraphics[width=\textwidth,trim=5cm 5cm 10cm 10cm, clip]{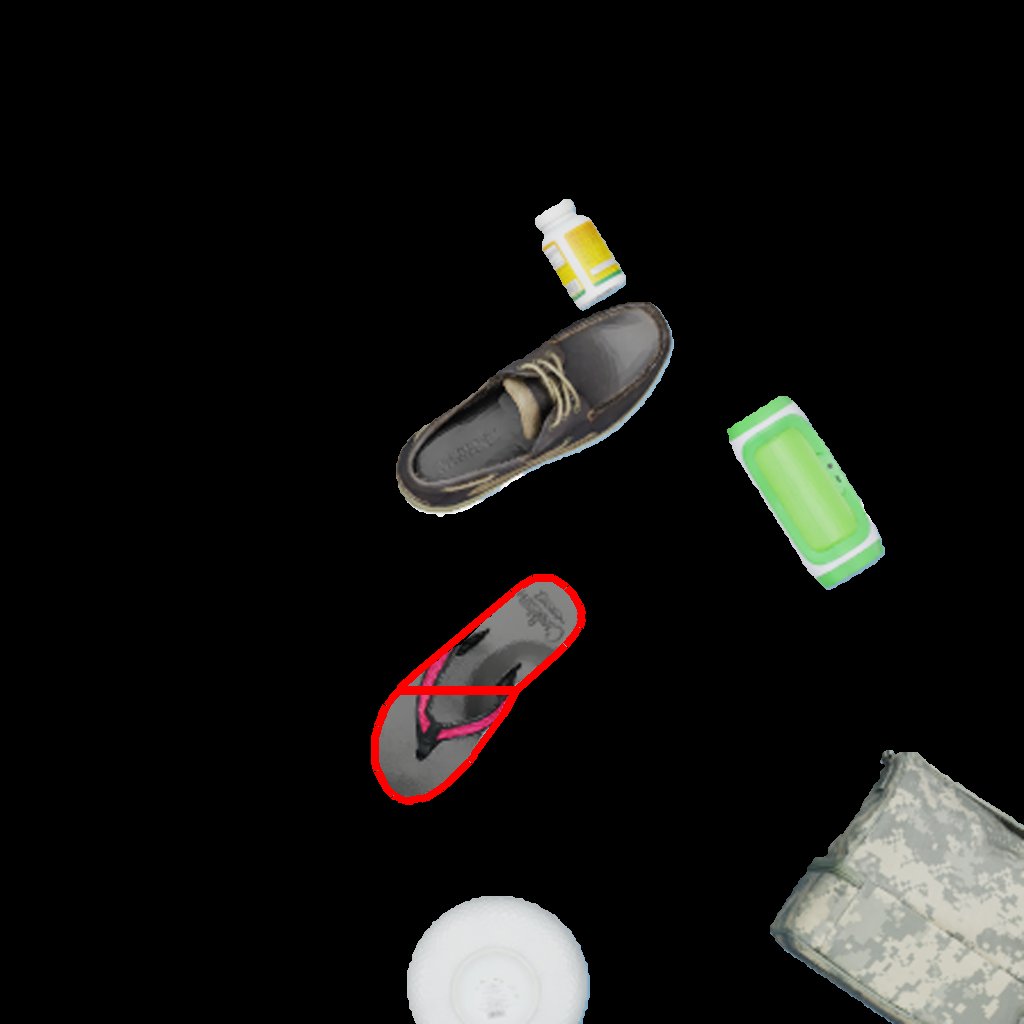}
        \caption*{}
    \end{subfigure}

    \caption{Bad examples of part segmentation by Shapegrasp. Objects are split too aggressively or it is too relaxed}
    \label{fig:shapegrasp_bad_examples}
\end{figure}

\begin{figure}
    \centering
    \begin{subfigure}[b]{0.3\textwidth}
        \centering
        \includegraphics[width=\textwidth, trim=7.5cm 7.5cm 7.5cm 7.5cm, clip]{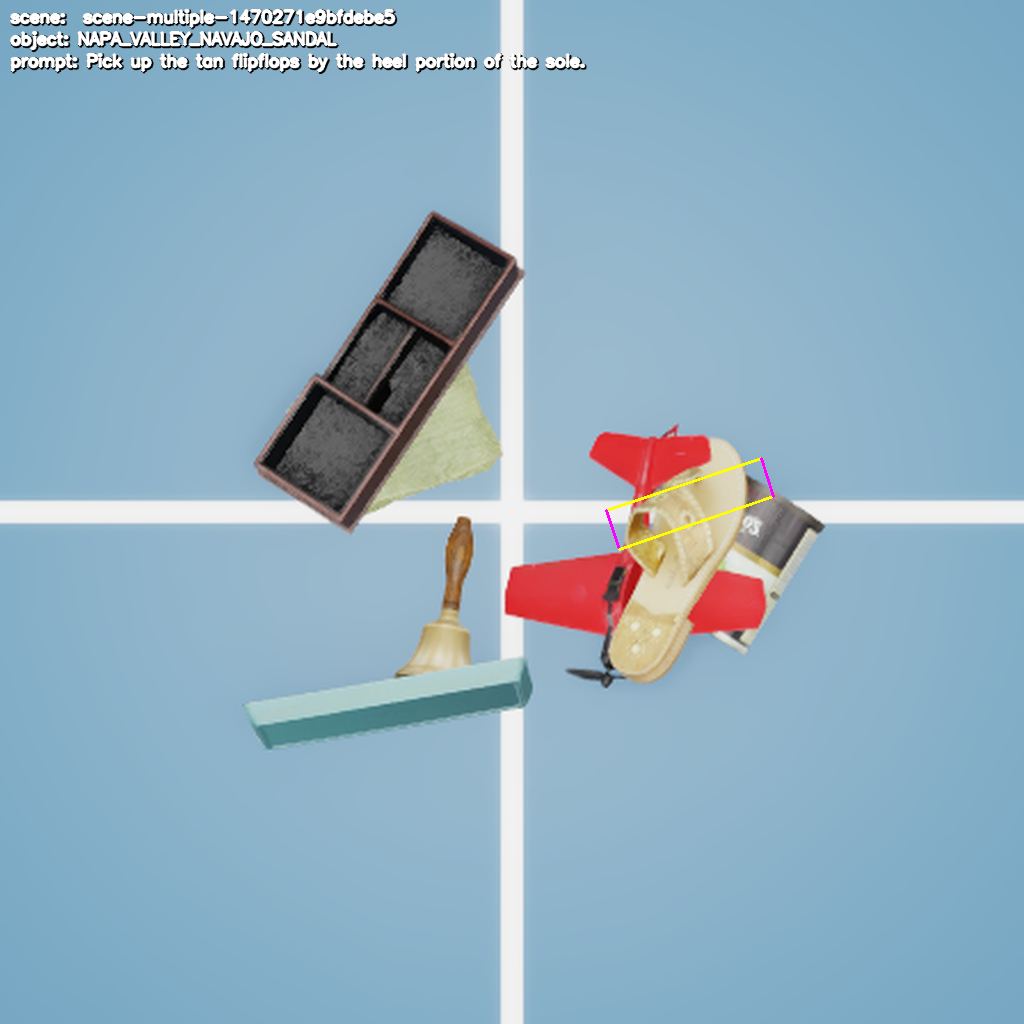}
        \captionsetup{justification=centering}
        \caption*{"Pickup the tan flipflop by the heel portion of the sole"}
    \end{subfigure}
    \hfill
    \begin{subfigure}[b]{0.3\textwidth}
        \centering
        \includegraphics[width=\textwidth, trim=0 5cm 10cm 5cm, clip]{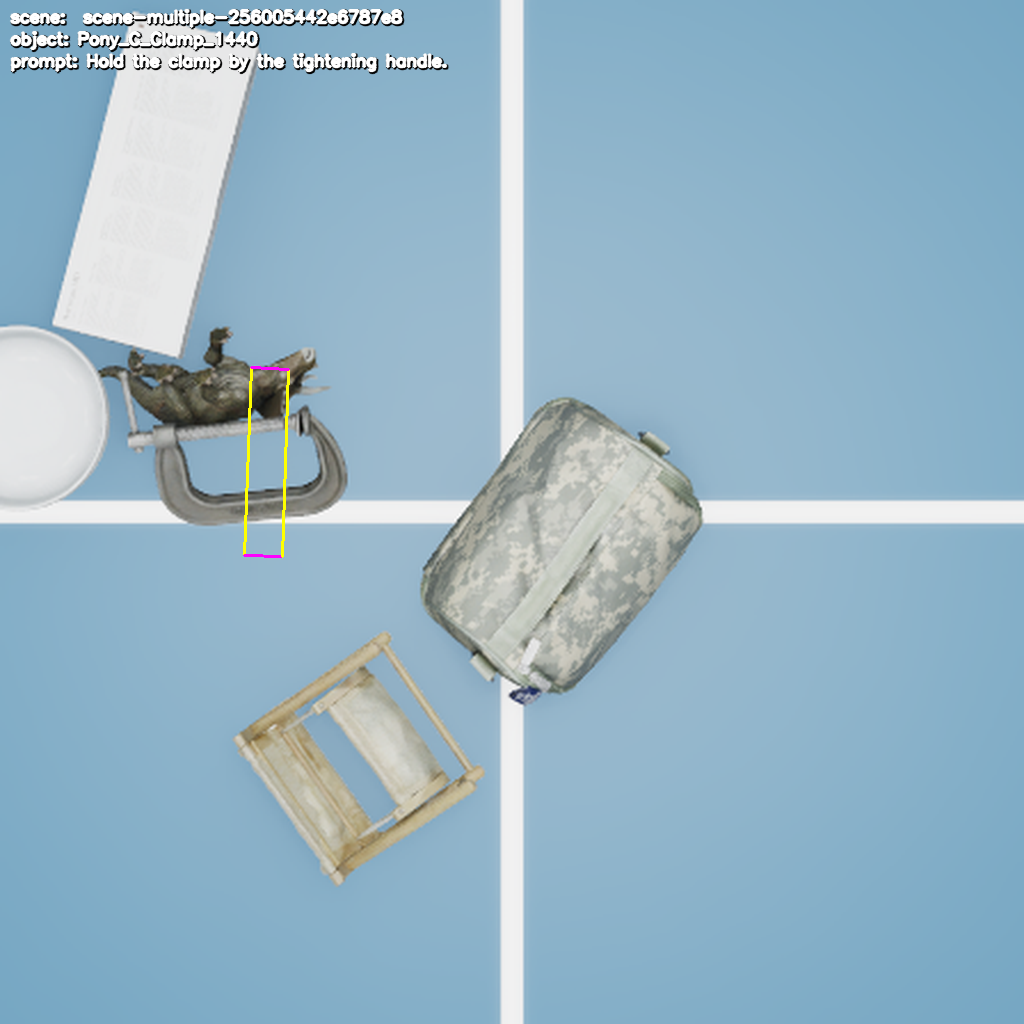}
        \captionsetup{justification=centering}
        \caption*{"Hold the clamp by the tightening handle"}
    \end{subfigure}
    \hfill
    \begin{subfigure}[b]{0.3\textwidth}
        \centering
        \includegraphics[width=\textwidth,trim=0 12.5cm 15cm 2.5cm, clip]{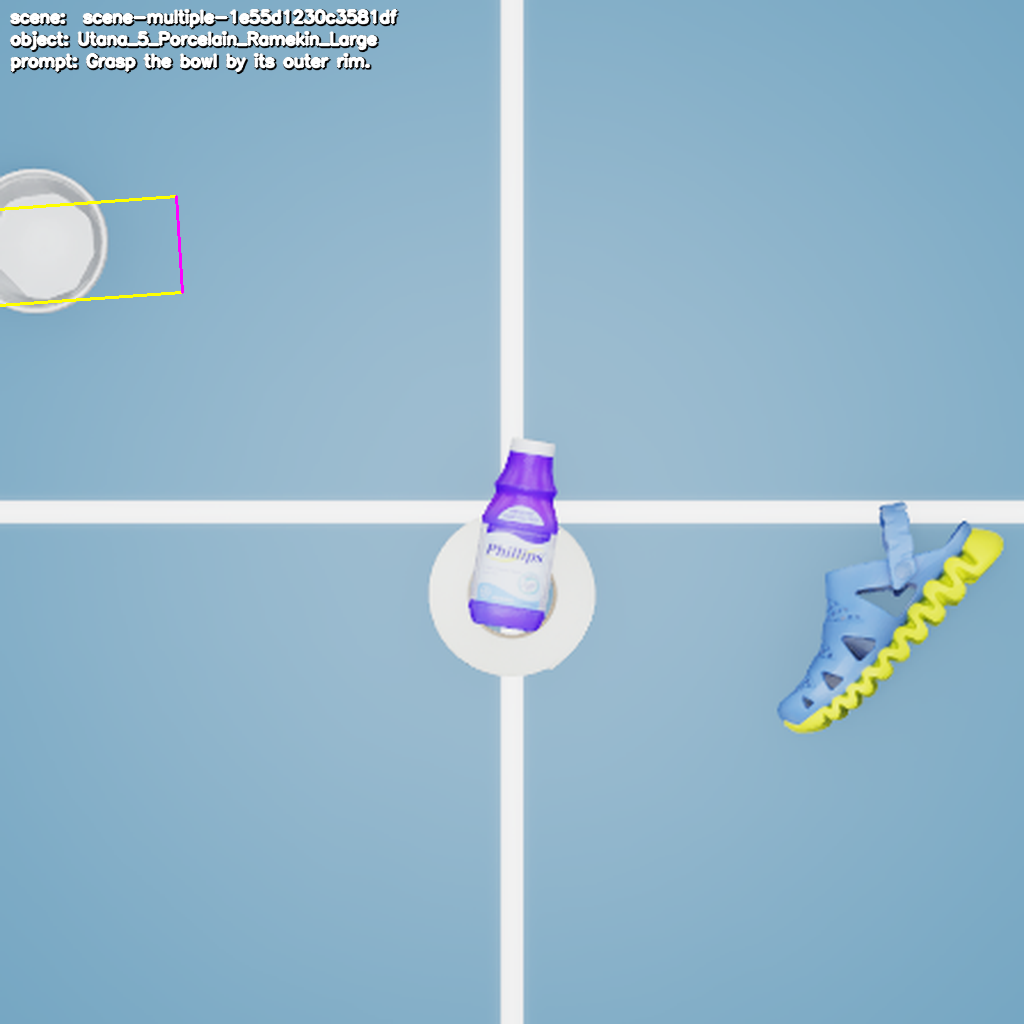}
        \captionsetup{justification=centering}
        \caption*{"Grab the white bowl by its outer rim"}
    \end{subfigure}

    \caption{Imprecise grasp locations from GraspMAS. First two examples show wrong part being grasped, and the third one is too imprecise with its location.}
    \label{fig:graspmas_examples}
\end{figure}

\paragraph{ShapeGrasp:} Shapegrasp works by decomposing the objects in the input image and then selecting the part of the object that best matches with the input prompt that allows for the grasp to be successful. \Cref{fig:shapegrasp_good_examples} and \Cref{fig:shapegrasp_bad_examples} highlight some of the good and bad examples of object part decomposition. The method performs well when the object are non-convex and can be decomposed into smaller convex parts (ex. the red toy plane and C-Clamp). For objects which are already mostly convex, the decomposition is either too aggressive like in the case of green speaker or too relaxed like in the brown shoe where the actual grasp prediction can be too imprecise.
\paragraph{GraspMAS:} Failures for this baseline were a mix of recognizing wrong objects in the scenes or being too imprecise with the grasp location resulting in grabbing the wrong part of the object. Examples of such imprecise predictions are given in \Cref{fig:graspmas_examples}.

Further, as mentioned in \Cref{sec:eval_lang_cond}, both GraspMAS and ShapeGrasp use only a Bird's Eye View image of the scene. Identifying the correct part is only the first step; even within that localized region, the model must still evaluate numerous valid grasp locations and orientations across the full 3D geometry. This becomes even more difficult when the object geometries are highly complex in nature and are often occluded in the BEV image.

\subsection{Dataset Size Ablation}
\begin{figure}[htbp]
    \centering
    \includegraphics[width=0.9\linewidth]{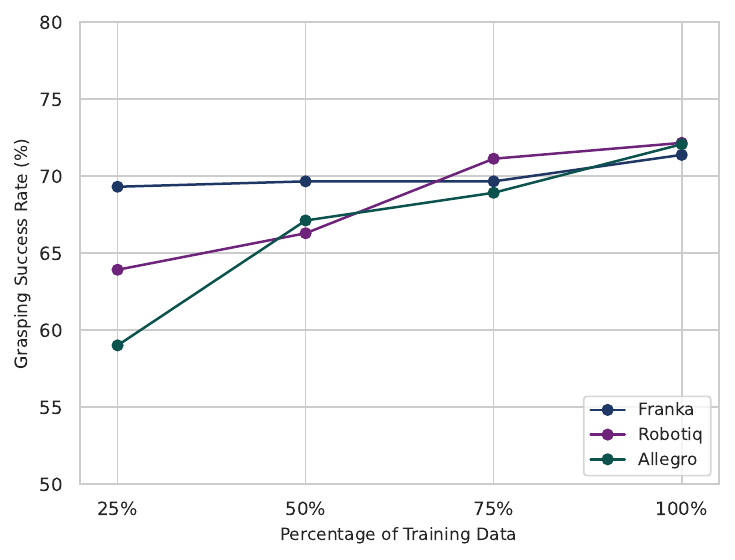}
    \caption{Success Rates with Varying Amount of Training Data}
    \label{fig:abl_data}
    \vspace{-15pt}
\end{figure}
For the Franka gripper, there is little effect of decreasing the dataset size, while the Robotiq 3-Finger and the Allegro Hand do benefit from the entire dataset. 
However, the marginal performance improvement decreases when more of the dataset is used.

\subsection{VLM Ablation}
\Cref{tab:abl_vlm_grippers} is an extended version of \Cref{tab:abl_vlm} with success rates per gripper.
The relative ranking of the three VLMs is the same for all grippers. 
\begin{table}[H]
    \centering
    \renewcommand{\arraystretch}{1.2} 
    \begin{tabular}{l||c|c|c}
    \hline
    \multicolumn{1}{c||}{\multirow{2}{*}{\textbf{Method}}} & \multicolumn{3}{c}{\textbf{Success (\%)}} \\ \cline{2-4}
    \multicolumn{1}{c||}{} & Franka Panda & Robotiq 3-Finger & Allegro \\ \hline \hline
    \rowcolor{uoftsecondaryblue_85_tint} GPT 5.4 & 37.7 & 41.47 & 48.65 \\
    Qwen 3.5 Flash~\cite{qwen35blog} & 46.79 & 51.85 & 64.63 \\
    \rowcolor{uoftsecondaryblue_85_tint} Gemini 3.1 Flash & \textbf{71.38} & \textbf{72.16} & \textbf{72.07} \\
    \hline
    \end{tabular}
    \caption{Success Rates using various VLMs for seed point prediction for all three grippers. Gemini 3.1 Flash outperforms others by a large margin.}
    \vspace{-15pt}
    \label{tab:abl_vlm_grippers}
\end{table}

\subsection{VLM Prompt}
\label{sec:vlm_prompt}

The prompt is sent as a chat-style request consisting of (i) a \texttt{role=system} message containing the system prompt (Box~1) and (ii) a single \texttt{role=user} message whose \texttt{content} is a sequence of alternating text and image parts (Box~2). For each scene $i$, the user message first appends a text segment of the form ``Scene number $i$: Object description: \textless USER\_INSTRUCTION\textgreater'', and then appends the corresponding image as an \texttt{image\_url} content part with a data URL of the form \texttt{data:image/jpeg;base64,<BASE64\_IMAGE\_i>}.

\begin{tcolorbox}[
    breakable,
    colback=gray!5!white, 
    colframe=gray!75!black,
    title={System Prompt (API message role = \texttt{system})},
]

You are a robotic manipulation specialist. Please help determine a stable grasp location for the specified object in the provided image of each scene. The robot gripper will attempt to grab the object at that exact location.

Express the desired grasp location as a pixel coordinate (y, x) normalized to the 0--1000 range.

Output the result as a JSON in the following format:

Example JSON result:
\begin{Verbatim}
{
  "scene_number": {
    "object": "object_description",
    "coords": [y, x],
    "explanation": "brief_justification"
  }
}
\end{Verbatim}

Critical Requirements:
\begin{enumerate}
  \item Ensure that the grasp location is free from collisions with surrounding objects.
  \item Ensure that the location is on the surface of the object as it will be projected onto the 3D point cloud of the scene. The location must be on the target object, where the gripper will grasp; it cannot be in empty space for hollow objects.
  \item Ensure that the gripper can encapsulate the object around the chosen grasp location. Grippers are standard sizes and cannot fully wrap around large objects. In such cases, edges or outcrops may be a better grasping location; do not attempt to grab the entire object body.
  \item Ensure that your output follows the specified JSON format.
\end{enumerate}
\end{tcolorbox}

\begin{tcolorbox}[
    breakable,
    colback=gray!5!white, colframe=gray!75!black,
    title={User Message Content (API message role = \texttt{user}), 3 scenes}
    ]
Scene number 1:

Object description: \textless USER\_INSTRUCTION\textgreater
\begin{Verbatim}
{
    "type":"image_url",
    "image_url":
            {
                "url":"data:image/jpeg;base64,<BASE64_IMAGE_3>"
            }
}\end{Verbatim}

Scene number 2:

Object description: \textless USER\_INSTRUCTION\textgreater
\begin{Verbatim}
{
    "type":"image_url",
    "image_url":
            {
                "url":"data:image/jpeg;base64,<BASE64_IMAGE_3>"
            }
}\end{Verbatim}

\end{tcolorbox}

\subsection{Real World Individual Item Success Rates}
The model was able to predict feasible grasps for all objects. The main cause of failures was that the gripper was too far from the object, causing it to close without actually grasping the object. This was amplified by the grasp sometimes being very angled. 

\begin{table}[h]  
\centering  
\begin{tabular}{|l|c|}  
\hline  
\textbf{Object} & \textbf{Success Rate} \\  
\hline  
Pink shark plushy & 7/10 \\  
Red plush dog & 9/10 \\  
Black cardboard box & 8/10 \\  
Toilet paper roll & 9/10 \\  
Paper towel roll & 9/10 \\  
Large rubber duck & 6/10 \\  
Small rubber duck & 9/10 \\  
Blue sneaker & 7/10 \\  
Gray shoe & 8/10 \\  
Blue towel & 8/10 \\  
Bike helmet & 8/10 \\  
Brown bucket & 7/10 \\  
Blue mug & 8/10 \\  
Green bottle & 6/10 \\  
Multicolor broom & 8/10 \\  
\hline  
\end{tabular}  
\caption{Grasp success rate for individual real-world objects.}  
\label{tab:real_performance}  
\end{table}
\end{document}